\documentclass[12pt,a4paper]{article}

\usepackage[british]{babel}

\usepackage[a4paper,top=2cm,bottom=2cm,left=2.5cm,right=2.5cm,marginparwidth=1.75cm]{geometry}

\usepackage[style=apa, backend=biber]{biblatex} 
\DeclareLanguageMapping{british}{british-apa} 
\DeclareFieldFormat[article]{volume}{\apanum{#1}} 

\usepackage{amsmath}
\usepackage{graphicx}
\usepackage[colorlinks=true, allcolors=blue]{hyperref}
\usepackage{hyperref}
\usepackage[title]{appendix}
\usepackage{mathrsfs}
\usepackage{amsfonts}
\usepackage{booktabs} 
\usepackage{caption}  
\usepackage{threeparttable} 
\usepackage{algorithm}
\usepackage{algorithmicx}
\usepackage{algpseudocode}
\usepackage{listings}
\usepackage{enumitem}
\usepackage{chngcntr}
\usepackage{booktabs}
\usepackage{lipsum}
\usepackage{subcaption}
\usepackage{authblk}
\usepackage[T1]{fontenc}    
\usepackage{csquotes}       
\usepackage{diagbox}
\usepackage{comment}
\usepackage{url}
\usepackage{lmodern}

\usepackage[justification=raggedright, singlelinecheck=false]{caption}
\usepackage{helvet}  
\usepackage{lmodern}  

\usepackage{setspace}
\usepackage{titlesec}
\titleformat{\section} 
  {\normalfont\Large\bfseries}{\thesection.}{1em}{}
  
\usepackage{lineno} 

\rightlinenumbers 

\linenumbers 

\usepackage{float}   
\usepackage{caption}


\date{}  

\begin{document}


\begin{center}
{\LARGE A Unified Framework for Kinematic Simulation of Rigid Foldable Structures} \\[5ex]

{\large Dongwook Kwak} \\
{\small
Department of Mechanical Engineering, Seoul National University, Seoul, Republic of Korea \\
ume1838@snu.ac.kr
} \\[1.2em]

{\large Geonhee Cho} \\
{\small
Department of Mechanical Engineering, Seoul National University, Seoul, Republic of Korea \\
gunhee236@snu.ac.kr
} \\[1.2em]

{\large Jiook Chung} \\
{\small
Department of Mechanical Engineering, Seoul National University, Seoul, Republic of Korea \\
jiwook2000@snu.ac.kr
}  \\[1.2em]

{\large Jinkyu Yang} \\
{\small
Department of Mechanical Engineering, Seoul National University, Seoul, Republic of Korea \\
jkyang11@snu.ac.kr
}
\end{center}

\begin{abstract}
Origami-inspired structures with rigid panels now span thick, kirigami, and multi-sheet realizations, making unified kinematic analysis essential. Yet a general method that consolidates their loop constraints has been lacking. We present an automated approach that generates the Pfaffian constraint matrix for arbitrary rigid foldable structures (RFS). From a minimally extended data schema, the tool constructs the facet–hinge graph, extracts a minimum cycle basis that captures all constraints, and—via screw theory— assembles a velocity-level constraint matrix that encodes coupled rotation-translation loop closure. The framework computes and visualizes deploy/fold motions across diverse RFS, while eliminating tedious, error-prone constraint calculation.
\end{abstract}

\textbf{Keywords}: rigid origami, kinematic analysis, loop closure constraints, rigid-foldable structures, screw theory, facet-hinge graph



\section{Introduction}
Origami-inspired structures have attracted broad attention in engineering owing to their unique combination of compact stowage, large deployability, and structural reconfigurability.  
Such properties have enabled diverse applications—from deployable space structures \parencite{zhang2023,sun2024} to reconfigurable robots \parencite{lerner2024,chen2022} and adaptive architectural systems \parencite{zhu2024}—where compactness and controlled motion are crucial.  

Among these, \textit{rigid origami}—where each panel moves as a rigid body without bending or stretching—has become a key foundation for engineering origami design.  
This concept naturally extends beyond the classical thin-sheet origami to a broader class of \textit{rigid foldable structures} (RFSs). These include thick-panel implementations, kirigami with perforations, and multi-sheet assemblies.  
These developments have diversified the mechanical and functional landscape of origami-inspired systems, while also amplifying the need for general kinematic analysis frameworks that can accurately capture the motion compatibility among rigid facets connected by hinges.

In such structures, the network of hinges connecting adjacent facets forms closed loops, and each loop imposes geometric closure constraints that restrict the feasible folding angles of the connected panels.  
Determining the folding angles that satisfy these constraints—or, given a configuration, computing the admissible motion directions—constitutes the core problem of rigid-foldable kinematics.  
This task is inherently challenging, as the constraints are nonlinear and strongly coupled even for relatively simple patterns.

To address this challenge, previous studies have attempted to compute feasible folding motions either analytically or numerically.  
Analytic approaches can be largely categorized into two groups.  
The first group derives explicit trigonometric relationships among folding angles within a unit cell, such as in symmetric waterbomb origami \parencite{chen2016}, Miura and stacked Miura origami \parencite{schenk2013}, and Tachi–Miura Polyhedron (TMP) patterns \parencite{yasuda2013}.  
The second group adopts a rigid-body formulation based on the Denavit–Hartenberg (DH) convention. Each panel is manually assigned a local frame, and the $4\times4$ transformation matrices between adjacent panels are computed to satisfy loop closure.  
This method has been applied to thick-panel origami and kirigami to evaluate feasible joint angles that yield identity transformations \parencite{chen2015,yang2022,wang2024}.  
While these approaches offer analytic insight into specific geometries, their extension to general, multi-loop configurations remains intractable due to the high algebraic complexity.

Numerical methods have aimed to generalize the analysis by constructing velocity-level constraint matrices.  
\textcite{tachi2009} developed a numerical solver, which efficiently determines foldable configurations but focuses mainly on thin, non-perforated origami where only rotational compatibility is enforced.  
More recently, \textcite{suto2023} introduced the \textit{CRANE} platform, enabling interactive kinematic simulations for arbitrary crease patterns. However, it also assumes single-layer origami and does not account for translational coupling or inter-layer constraints.

Finite-element–based simulation approaches, including the bar-and-hinge model \parencite{liu2017} and \textit{Origami Simulator} \parencite{ghassaei2018}, can handle large-scale folding problems at low computational cost.  
Nevertheless, they allow facet deformation for numerical stability, and thus cannot strictly preserve rigid-folding kinematics—such as singularities, constraint satisfaction, and degrees of freedom—required for accurate analysis of rigid foldable structures.  
Comprehensive survey of kinematic modeling strategies and practical formulations of loop-closure constraints in origami and related systems can be founded in the recent review by \textcite{zhu2022}.

This study presents a unified computational framework for the kinematic analysis of RFSs.  
An RFS is formulated as a spatial linkage composed of rigid facets and revolute joints.  
We introduce a minimal yet general geometric data schema to represent arbitrary fold patterns and their inter-sheet connections.  
Using a graph-based approach, internal loops are automatically identified to capture independent closure constraints,  
while the screw-theoretic formulation expresses coupled rotational and translational motions in a unified matrix form.  
Together, these components provide a compact and scalable foundation for analyzing diverse rigid-foldable structures.

\section{Methodology}

\subsection{Overview of the Framework}
\begin{figure}[!ht]
\centering
\includegraphics[width=1\linewidth]{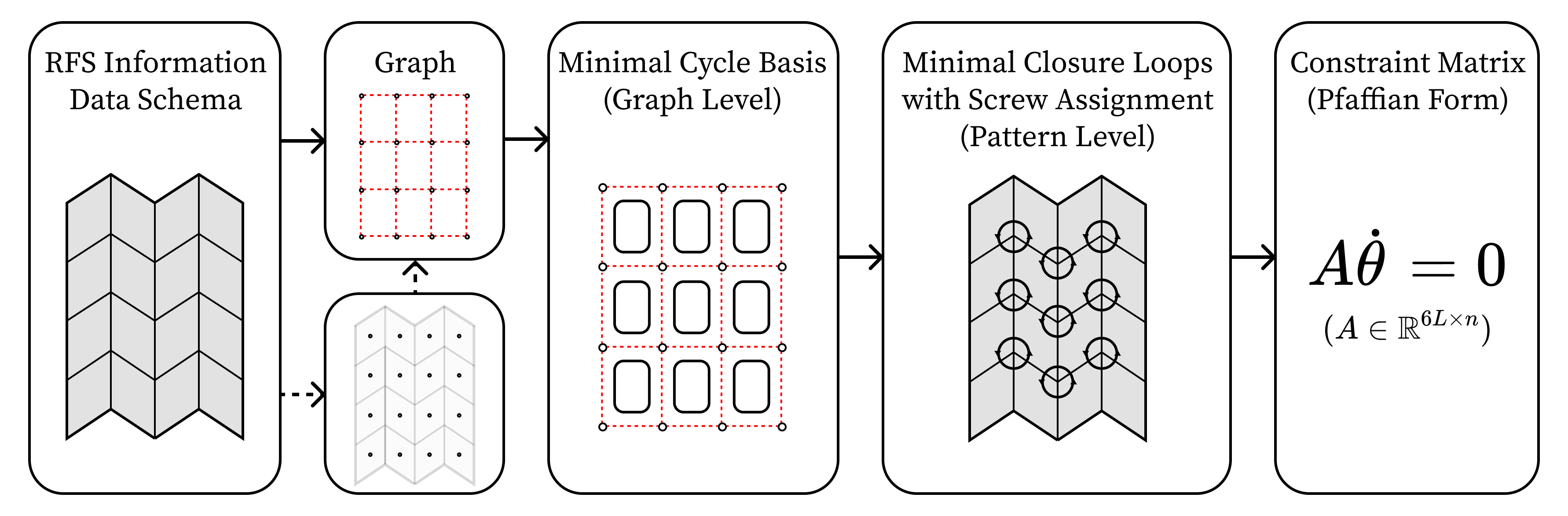}
\caption{\label{fig:flowchart}Overview of the proposed computational pipeline for constructing the Pfaffian constraint matrix of an RFS.}
\end{figure}

The ultimate goal of the proposed framework is to construct the Pfaffian constraint matrix of a given RFS that encodes the velocity-level motion constraints among the hinges.  
This matrix should capture all loop-closure relationships of the structure, ensuring that the folding motion satisfies both rotational and translational compatibility across connected facets.

Figure~\ref{fig:flowchart} illustrates the overall computational pipeline.  
The framework proceeds through four major stages: (i) data schema of RFS, (ii) graph construction and closed-loop extraction, (iii) screw assignment, and (iv) Pfaffian matrix assembly.  
Each stage progressively develops geometric information into a complete mathematical representation of rigid-folding motion.

We first define a minimal yet general representation of RFS that provides an intuitive input for users while preserving all geometric information necessary for kinematic modeling.    
This unified representation allows both single-layer origami and multi-layer configurations to be handled consistently.
From this input, a facet–hinge graph is built to automatically identify independent closure loops. 
Screw axes are then assigned consistently across these loops, and all resulting constraints are assembled into a compact velocity-level formulation.  
Detailed explanations of each stage, including how the data schema operates, are provided in the following sections.

\subsection{\label{sec:data schema} Data Schema of Rigid Foldable Structures}
\begin{figure}[!ht]
\centering
\includegraphics[width=1\linewidth]{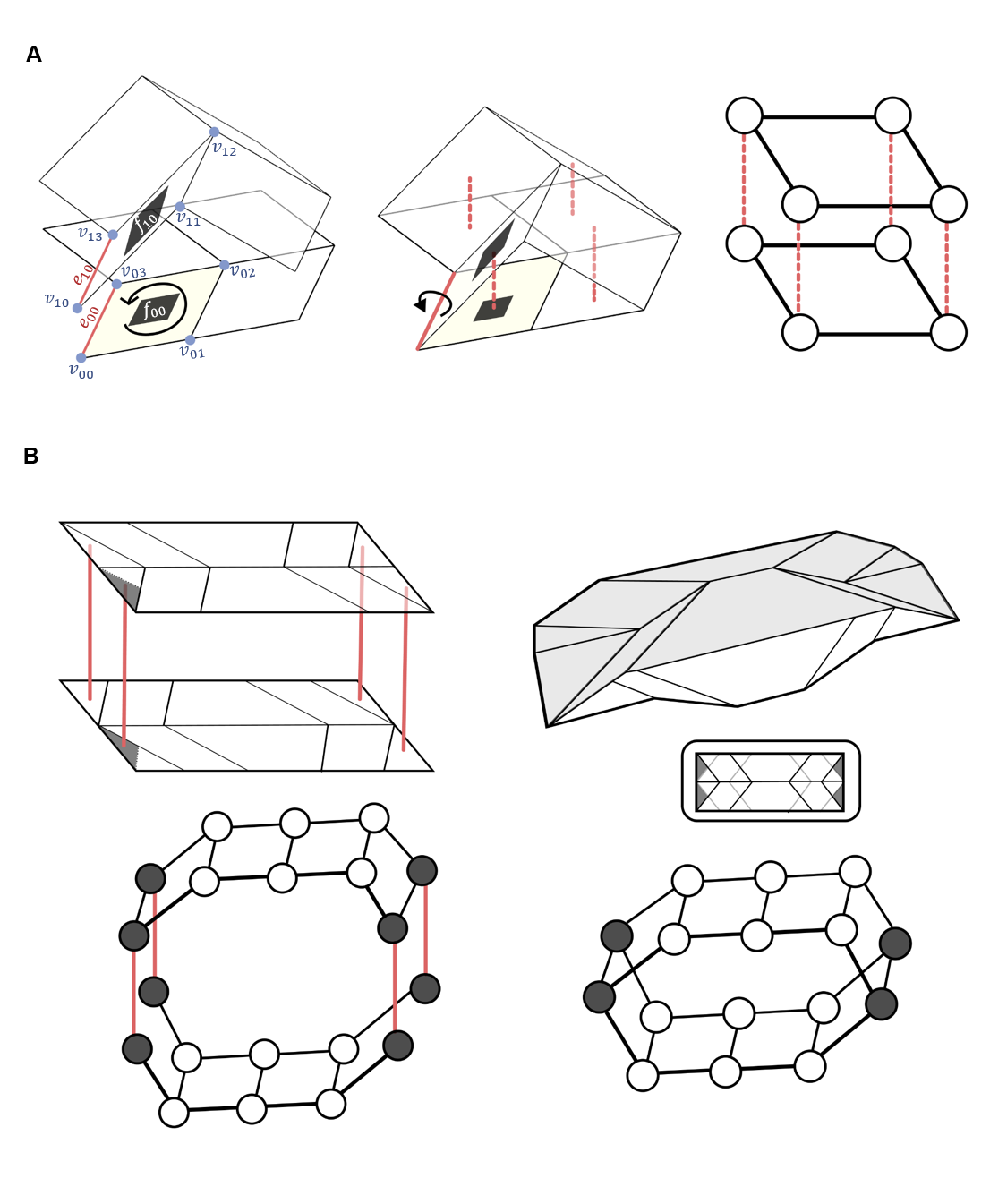}
\caption{\label{fig:data_example}
Examples of sheet-wise connections and their graph representation in the proposed data schema.  
(A) Hinging-type connection introducing a rotational degree of freedom in a stacked Miura origami (SMO) unit cell.  
(B) Soldering-type connection constraining relative motion in a Tachi--Miura polyhedron (TMP) unit cell.
In the SMO case, hinging adds graph edges, whereas in the TMP case, soldering merges nodes into a single one, resulting in distinct graph representations.}
\end{figure}


We introduce a data schema that explicitly encodes both intra-sheet and inter-sheet relationships to systematically represent the geometric and kinematic configuration of multi-sheet RFSs.  
This schema provides a unified and modular description of all geometric entities—vertices, pattern edges, and facets—as well as the sheet-wise connections between them, forming the basis for graph construction and constraint analysis.


Figure~\ref{fig:data_example}A shows an example of the notations used in this study. Each geometric element is indexed by its sheet number $i$ and its local index $j$:  
$v_{ij}$, $e_{ij}$, and $f_{ij}$ denote the $j^{\text{th}}$ vertex, pattern edge, and facet on the $i^{\text{th}}$ sheet, respectively.  
A vertex $v_{ij}$ stores its 3D coordinates, e.g., $v_{00} = [x_{00}, y_{00}, z_{00}]$.  
A pattern edge $e_{ij}$ represents a pair of vertex indices—for example, $e_{00} = [v_{00}, v_{03}]$ and $e_{10} = [v_{10}, v_{13}]$.  
The term \emph{pattern edge} is used here to distinguish these geometric edges from the graph edges that will be introduced later.  
Each facet $f_{ij}$ is defined as an ordered list of vertex indices on sheet $i$; for instance, $f_{00} = [v_{00}, v_{01}, v_{02}, v_{03}]$.  
Across different sheets, sheet-wise connections can be manually specified by assigning two facet indices and a connection type—for example, $c_1 = [f_{00}, f_{10}, h]$ for a hinging-type connection or $c_1 = [f_{00}, f_{10}, s]$ for a soldering-type connection.  
Representative examples of the hinging- and soldering-type connections and illustrated in Figure~\ref{fig:data_example}A and Figure~\ref{fig:data_example}B, respectively.

The overall schema is formally defined as follows:

\begin{itemize}
\item \textbf{Vertices:}
$V = \left[ [v_{00},\dots,v_{0N_{V,1}}], \dots, [v_{m0},\dots,v_{mN_{V,m}}] \right]$,  
where each $v_{ij}\in\mathbb{R}^3$ represents the 3D coordinate of the $j$-th vertex on the $i$-th sheet.  
$N_{V,i}+1$ denotes the number of vertices on sheet $i$.

\item \textbf{Pattern edges:}
$E = \left[ [e_{00},\dots,e_{0N_{E,1}}], \dots, [e_{m0},\dots,e_{mN_{E,m}}] \right]$,  
where each $e_{ij}\in\mathbb{Z}_{\ge 0}^{\,2}$ specifies a pair of vertex indices forming a pattern edge on sheet $i$.  
$N_{E,i}+1$ is the number of pattern edges on sheet $i$.

\item \textbf{Facets:}
$F = \left[ [f_{00},\dots,f_{0N_{F,1}}], \dots, [f_{m0},\dots,f_{mN_{F,m}}] \right]$,  
where $f_{ij}\in\mathbb{Z}_{\ge 0}^{\,\ell(i, j)}$ lists the vertex indices defining the $j$-th facet on sheet $i$ (with $\ell(i, j)$ vertices).  
$N_{F,i}+1$ is the number of facets on sheet $i$.

\item \textbf{Sheet-wise connections:}
$C = [c_1,\dots,c_{N_C}]$, where $c_\alpha = [f, f, x]$ specifies two connected facets (necessarily on different sheets) and a connection type $x\in\{h,s\}$.  
Here $x=h$ (\emph{hinging}) introduces a rotational degree of freedom along their shared boundary,  
while $x=s$ (\emph{soldering}) constrains the two facets to move as one rigid body.  
$N_C$ denotes the total number of inter-sheet connections.
\end{itemize}

Each sheet designates its first facet as a \emph{seed facet}, whose boundary orientation must be explicitly defined, either counterclockwise or clockwise.
This orientation determines the facet normal and thereby defines the mountain–valley configuration of intra-sheet creases.  
Once the seed facet is defined, all other facets within the same sheet are automatically aligned through shared pattern edges to ensure consistent orientation across the sheet. The resulting uniform facet orientation forms the basis for the consistent screw-direction assignment discussed in Section~\ref{sec:screw_assign},  
while additional conventions—such as connection grouping and the formal definition of a sheet—are summarized in Appendix~\ref{appendix:convention}.

Conceptually, one can think of this schema as a layer-wise extension of the widely used $(V,E,F)$ representation,  
augmented with sheet indices and connection attributes to handle multi-layer RFSs.  
By tagging all geometric elements with sheet indices and explicitly recording inter-layer connections,  
the proposed schema generalizes the conventional single-sheet representation to handle multi-layer configurations in a unified manner.  
This foundation allows the subsequent algorithms to be directly applied to a broad range of RFSs.

\subsection{\label{sec:graph_rep} Facet-Hinge Graph and Loop Extraction}
\begin{figure}[!htbp]
\centering
\includegraphics[width=1\linewidth]{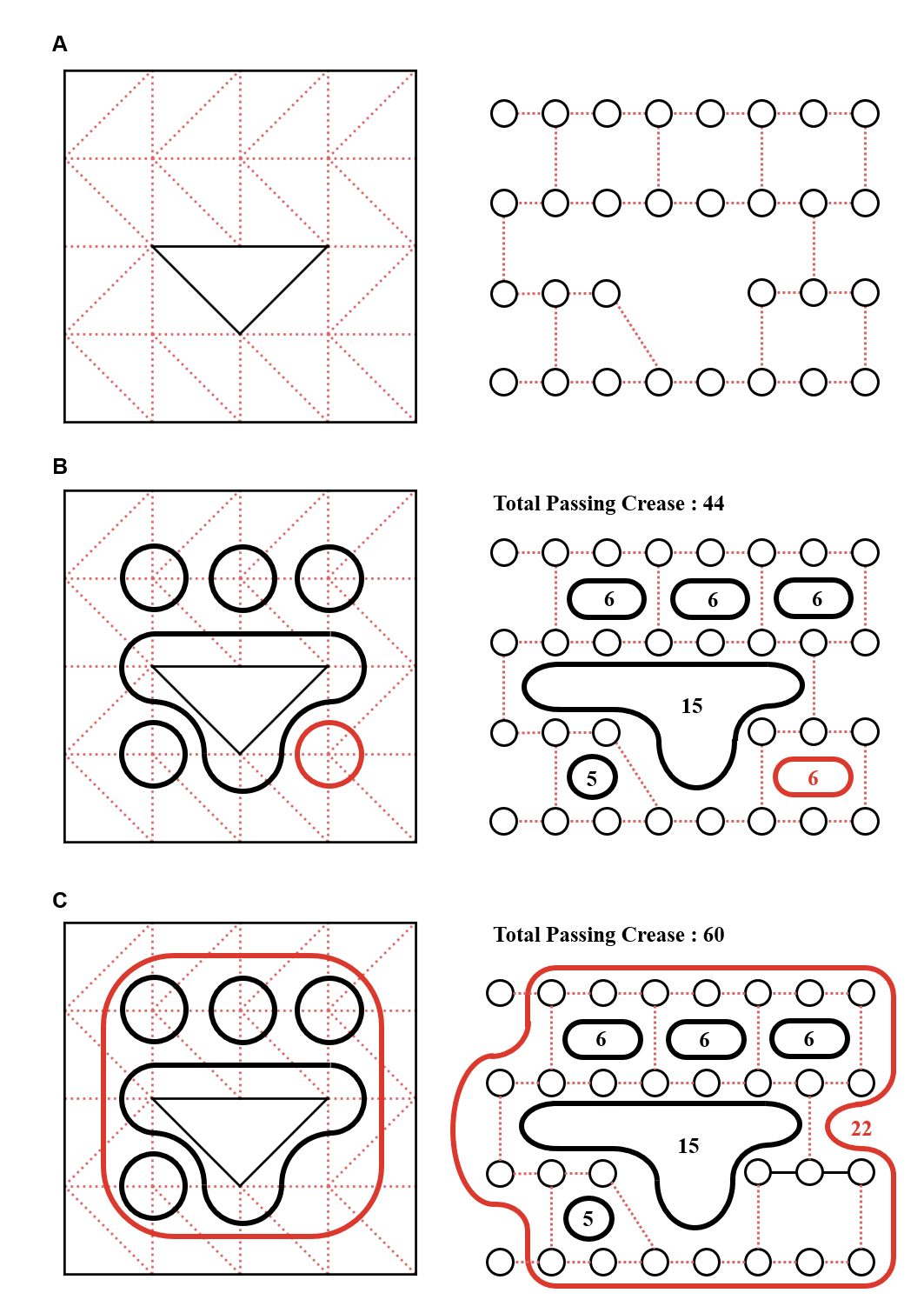}
\caption{\label{fig:graph_origami_analogy}
Closure loops in patterns and cycles in graphs are in one-to-one correspondence (left: pattern, right: graph).  
(A) Facet–hinge graph representation of a rigid-foldable kirigami pattern.
(B) A complete and minimum set of closure loops obtained via the minimum cycle basis.  
(C) A complete but non-minimum loop set that contains redundant cycles.}
\end{figure}

Starting from the data schema, we construct a \emph{facet–hinge graph} that captures the connectivity relationships among facets.  
Each facet becomes a \emph{graph node}, and intra-sheet pattern edges lying in the sheet interior serve as \emph{graph edges}, whereas boundary edges are excluded unless they participate in inter-sheet hinging connections.  
Inter-sheet connections of type $h$ (hinging) introduce new graph edges, while those of type $s$ (soldering) merge the connected nodes into a single node, producing distinct graph representations for the two connection types (Figure~\ref{fig:data_example}).  

The resulting facet–hinge graph typically contains many cycles, each corresponding to a closure loop in the RFS pattern.  
To obtain a complete and non-redundant set of such loops, we compute a \emph{minimum cycle basis} (MCB) \parencite{Horton1987}.  
The MCB provides the smallest collection of independent cycles from which all other cycles can be formed, corresponding exactly to the independent closure loops.  
Because the MCB minimizes the total loop length, each loop involves fewer hinge traversals, reducing the propagation of numerical error when the kinematic constraints are later assembled into the global constraint matrix. 

Figure~\ref{fig:graph_origami_analogy} illustrate the one-to-one correspondence between pattern-level closure loops and graph cycles.  
Each thick black loop in Figure~\ref{fig:graph_origami_analogy}B and \ref{fig:graph_origami_analogy}C denotes a closure loop in the physical pattern, and the matching cycle on the right shows its graph-level representation. 
The red loops highlight the distinction between a minimal cycle basis and a non-minimal set: Figure~\ref{fig:graph_origami_analogy}B contains a complete and minimal collection of loops, whereas Figure~\ref{fig:graph_origami_analogy}C includes an additional red loop that renders the set non-minimal.  
This redundant cycle can be expressed as a linear combination of the loops in Figure~\ref{fig:graph_origami_analogy}B, confirming that it introduces no new independent constraint.
The numbers placed at the center of each cycle indicate the number of creases traversed by that loop.  
Summing these values yields 44 crease traversals in the minimum case and 60 in the redundant case, demonstrating that non-minimal loop sets impose unnecessary hinge evaluations during constraint propagation.

The graph-based formulation offers three concrete benefits:  
(i) closure loops and graph cycles are directly related, providing a purely combinatorial handle on constraint selection;  
(ii) the MCB guarantees minimality under a chosen edge weight without ad hoc bookkeeping; and  
(iii) once the graph is built, loop extraction is automatic and reproducible across large, multi-layer patterns.

Based on the defined data schema, the facet–hinge graph is constructed automatically by comparing vertex indices and vertex coordinates across both intra- and inter-sheet facets.  
This procedure also reveals joints corresponding to hinging-type connections.  
Further methodological details are provided in Appendix~\ref{appendix:graph_construction}.



\subsection{\label{sec:screw_assign} Screw Assignment}
\begin{figure}[!ht]
\centering
\includegraphics[width=1\linewidth]{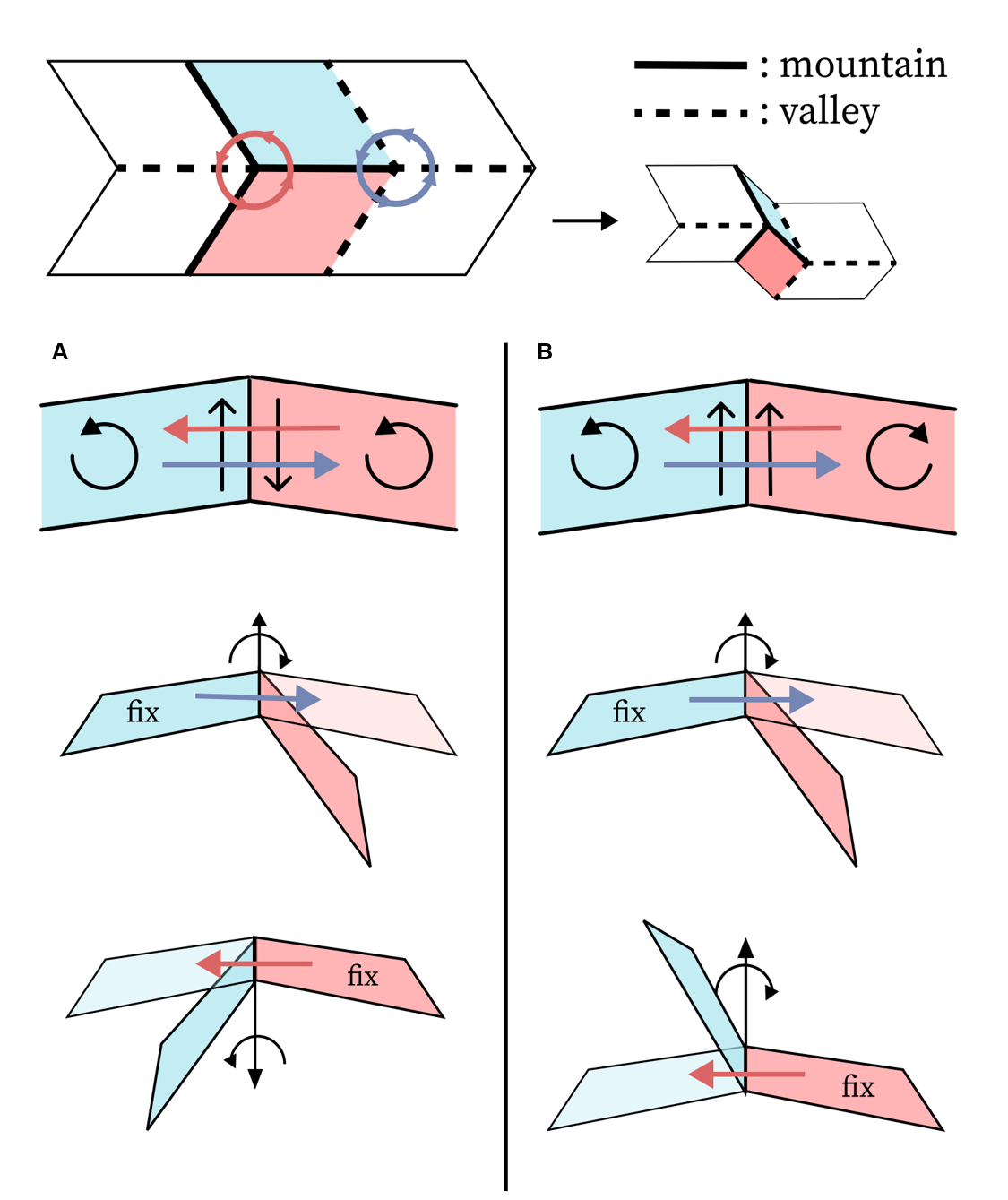}
\caption{\label{fig:screw_alignment}
Schematic illustration of screw-axis orientation consistency and its influence on motion continuity.
(A) Consistent facet orientation yields a uniform hinge motion.
(B) Reversed orientation causes conflicting screw directions and motion discontinuity.}
\end{figure}

The assignment of screw axes determines the rotation direction and polarity of fold motion across the structure.  
Building upon the orientation conventions defined in the data schema,  
all facet vertex loops on a given sheet share the same ordering.  
With this consistency, the hinge directions between adjacent facets can be aligned coherently,  
producing a continuous and conflict-free motion field across the sheet.
This geometric consistency enables the automatic assignment of hinge screws that generate continuous rigid-folding motions without sign conflicts. The overall concept of screw-axis alignment, governed by facet vertex loop ordering, and its effect on motion continuity is illustrated in Figure~\ref{fig:screw_alignment}.

The screw-axis assignment proceeds in three deterministic steps:

\begin{enumerate}
    \item \textbf{Seed facet initialization.}  
    A reference facet is first selected on each sheet, and its vertex order defines the positive orientation of all intra-sheet hinges.  
    A positive hinge rotation is defined such that, when viewed along the facet normal, it produces a mountain fold.
    
    \item \textbf{Loop-level unification.}  
    Within each minimal closure loop identified from the facet–hinge graph, the traversal direction is fixed according to the seed facet’s orientation.  
    Each hinge in the loop is assigned a space screw axis so that the rotational directions are consistent along the loop, avoiding any sign ambiguity at shared creases.
    
    \item \textbf{Global propagation.}  
    Because adjacent loops share at least one hinge, the assigned screw direction of a shared hinge is inherited by all loops containing it.  
    Propagating this rule through all loops yields a globally consistent screw-axis configuration without requiring manual correction.
\end{enumerate}

This procedure provides a method for determining all screw axes in arbitrary RFS, where no explicit local frames or parameter tuning are required.  
The screw assignment can also be derived directly from the information defined in the data schema,  
and the detailed computational pipeline is summarized in Appendix~\ref{appendix:screw_assign}.

\subsection{\label{sec:pfaffian} Constraint Matrix Assembly using Screw Theory}
From the facet–hinge graph constructed in the previous section,  
all graph edges corresponding to rotational joints are identified. 
The resulting set of screw axes forms the foundation for constructing the loop-level constraint matrices.
Collecting all revolute motions yields the set of \textit{hinge variables}  
$\theta \in \mathbb{R}^n$ used to describe the kinematic state of the structure.  

The holonomic loop-closure conditions among facets can be expressed purely in terms of the hinge angles $\theta$, making them \textit{scleronomic}, i.e., independent of time.  
Taking the time derivative gives their differential, or \textit{Pfaffian}, form:
\begin{equation}
A(\theta)\,\dot{\theta}=0,
\end{equation}
where $A(\theta)$ is the configuration-dependent \textit{Pfaffian constraint matrix}.  
The null space of $A(\theta)$ defines all admissible hinge velocities $\dot{\theta}$, and its rank determines the instantaneous degrees of freedom:
\begin{equation}
\mathrm{DoF}=n-\mathrm{rank}\left(A(\theta)\right).
\end{equation}

A nonzero DoF indicates the existence of feasible motion directions satisfying all closure constraints,  
allowing direct assessment of local foldability. In rare cases, such as singular configurations where the Jacobian of the hinge variables is ill-defined,  
the computed $A(\theta)$ may yield an apparent nonzero DoF even though no actual motion exists.  
Nonetheless, this formulation correctly predicts the true kinematic mobility for nearly all regular configurations encountered in practice.

In the following, we construct the constraint matrix using screw theory, which models each hinge as a screw motion between connected rigid facets. The mathematical background—covering screw representation, forward kinematics, and the space Jacobian—is summarized in the Appendix \ref{appendix:ScrewTheory}. Unlike DH parameterization, screw theory operates directly on geometric quantities such as axes and points, providing a coordinate-free and physically intuitive foundation for constraint formulation.

\begin{figure}[!ht]
\centering
\includegraphics[width=1\linewidth]{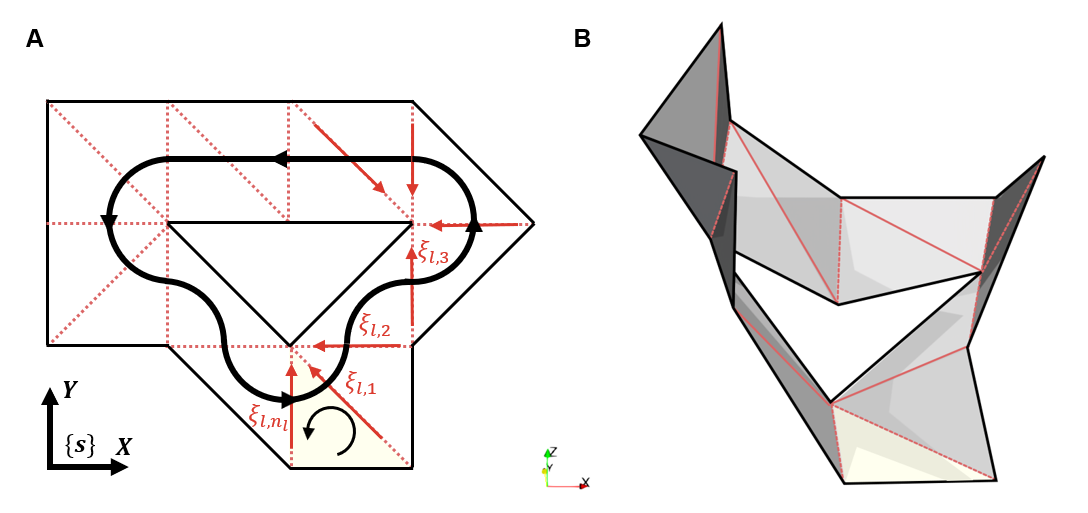}
\caption{\label{fig:loop_screw}
(A) Screw assignment and traversal direction of a single minimal closure loop.  
(B) Configuration after folding motion, where the loop closure constraint is satisfied.}
\end{figure}

Among the $L$ minimal closure loops identified from the facet–hinge graph, let the $l^{\text{th}}$ loop pass through $n_l$ crease lines.  
Each crease in the $l^{\text{th}}$ loop is assigned a local hinge variable $\rho_i$, used solely for internal constraint evaluation within that loop.  
The corresponding well-assigned space screw axis at the home position is denoted by $\xi_{l,i}$ $(i = 1, \dots, n_l)$ representing the geometric configuration of the $i^{\text{th}}$ hinge in its undeformed state (Figure~\ref{fig:loop_screw}).  

To derive the loop constraint, we virtually cut the loop and unfold it into an open chain.  
The last facet connected to the final hinge $\xi_{l,n_l}$ is assumed to coincide with the first facet in the loop, fixed to the global space frame $\{s\}$.  
This implies that the terminal and initial facet frames coincide in the home configuration, meaning that the home transformation matrix ($M_{s,n_l}$) of the end facet is the identity matrix.

After folding (i.e., rotating the hinges), the configuration of the end facet must still coincide with the space frame.  
Using the Product of Exponentials (PoE) formula, the loop closure condition is
\begin{equation}
T = e^{[\xi_{l,1}]\rho_1} e^{[\xi_{l,2}]\rho_2} \cdots e^{[\xi_{l,n_l}]\rho_{n_l}} M_{s,n_l} = I_{4}.
\end{equation}
Taking the time derivative yields
\begin{equation}
\dot{T} = \frac{\partial T}{\partial \rho_{1}} \dot{\rho}_{1} + \cdots + \frac{\partial T}{\partial \rho_{n_l}} \dot{\rho}_{n_l} = \mathbf{0}_{4\times4}.
\end{equation}
Defining $T_{j:k} = e^{[\xi_{l,j}]\rho_j} e^{[\xi_{l,j+1}]\rho_{j+1}} \cdots e^{[\xi_{l,k}]\rho_k}$,  
the partial derivative of $T$ with respect to $\rho_k$ becomes
\begin{equation}
\frac{\partial T}{\partial \rho_{k}} = T_{1:k-1}[\xi_{l,k}]T_{k:n_l}
=\begin{bmatrix}
	0 & -c_{l,k} & b_{l,k} & p_{l,k} \\
	c_{l,k} & 0 & -a_{l,k} & q_{l,k} \\
	-b_{l,k} & a_{l,k} & 0 & r_{l,k} \\
	0 & 0 & 0& 0
\end{bmatrix}.
\label{eq:Partial_derivative}
\end{equation}
Each derivative matrix $\frac{\partial T}{\partial \rho_k}$ belongs to the Lie algebra $se(3)$ and contains six independent components.
Accordingly, the 16 scalar equations implied by the zero-matrix condition reduce to six independent constraints, corresponding to the six degrees of freedom of a rigid body in 3D space.

By manipulating Equation~\ref{eq:Partial_derivative} and applying the definition of the adjoint operator (Equations~\ref{eq:Adjoint_start}--\ref{eq:Adjoint_end}), each derivative term can be compactly expressed as
\begin{equation}
\frac{\partial T}{\partial \rho_k} = T_{1:k-1} [\xi_{l,k}] T_{1:k-1}^{-1} \cdot T = [\mathrm{Ad}_{T_{1:k-1}}(\xi_{l,k})].
\end{equation}
This is identical in form to the $k^{\text{th}}$ column of the space Jacobian.
Hence, the loop closure constraint for the $l^{\text{th}}$ loop can be written using the \textit{loop space Jacobian}:
\begin{equation}
J_s^l = \begin{bmatrix}
\xi_{l,1} & \mathrm{Ad}_{T_{1:1}}(\xi_{l,2}) & \cdots & \mathrm{Ad}_{T_{1:n_l-1}}(\xi_{l,n_l})
\end{bmatrix},
\label{eq:Loop Closure Jacobian}
\end{equation}
where $T_{1:i}$ can be recursively evaluated through the exponential coordinates of motion.
This formulation provides a compact and efficient representation of the six independent closure constraints for each loop.

Re-indexing all loop Jacobians to the global hinge coordinate vector $\theta$ yields the configuration-dependent Pfaffian constraint matrix:
\begin{equation}
A(\theta)\dot{\theta} =
\begin{bmatrix} A_1 \\ A_2 \\ \vdots \\ A_{L} \end{bmatrix}\dot{\theta} = 0,
\qquad A(\theta)\in\mathbb{R}^{6L\times n}.
\end{equation}
where each $A_l$ corresponds to the loop Jacobian $J_s^l$ mapped to the appropriate global indices.
Through the use of screw-theoretic operators and adjoint transformations, the entire constraint formulation can thus be derived systematically and expressed in a unified coordinate.

In the case of non-perforated, single-sheet origami patterns, where all creases in a loop intersect at a common vertex,
the translational components of each loop constraint become linearly dependent on the rotational ones.
This special case can be rigorously derived within the same screw-theoretic framework and is summarized in Appendix~\ref{appendix:ScrewTheory2}.

\section{\label{sec:result} Results and Discussion}
Across all tested configurations, the proposed framework successfully generated valid constraint matrices and computed feasible folding angles for a wide variety of rigid-foldable structures, demonstrating robustness across diverse geometric complexities and structural configurations. Rather than focusing on solver performance alone, our open-source algorithm provides an integrated pipeline that takes pattern geometry encoded by the proposed data schema as input, computes equilibrium folding-angle trajectories by following prescribed crease neutral angles and stiffness distributions, and exports both numerical angle trajectories and corresponding visualizations within a single workflow.

\begin{figure}[!ht]
\centering
\includegraphics[width=1\linewidth]{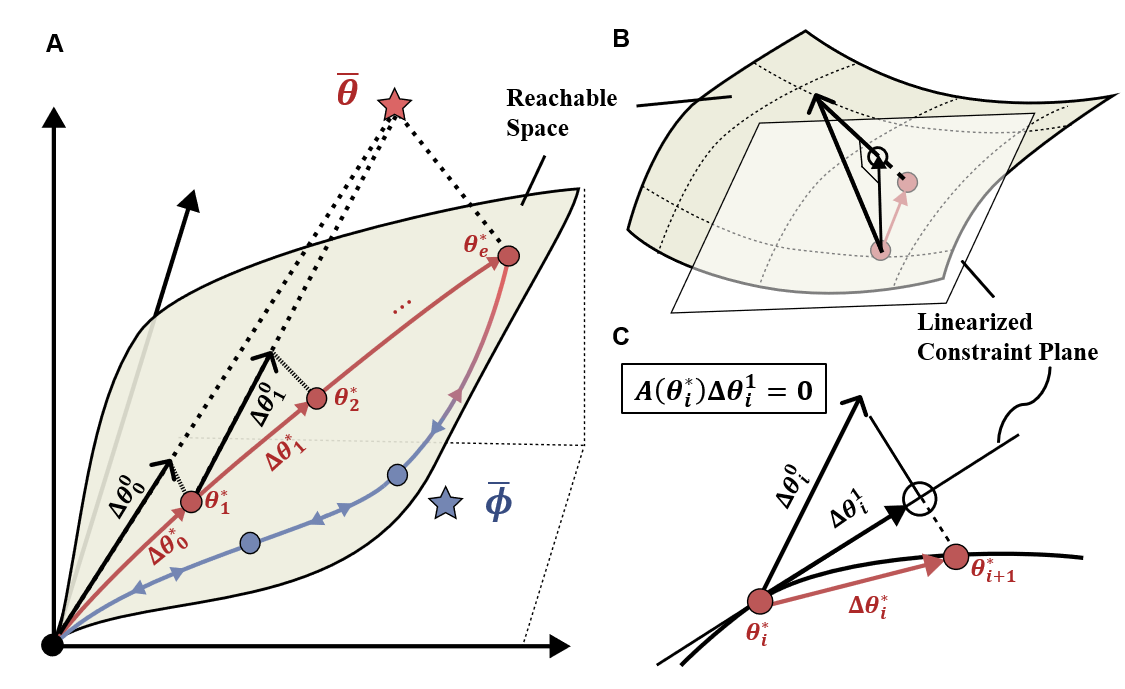}
\caption{\label{fig:newton_raphson}
Constraint-matrix–based feasible folding trajectory computation.
(A) Conceptual illustration of exploring feasible trajectories within the reachable space embedded in the $n$-dimensional hinge-variable space.
(B) Local update at a converged configuration: an unconstrained energetic update $\Delta\theta_i^0$ (open black arrow) is projected onto the linearized constraint plane (solid black line), yielding a constraint-consistent increment $\Delta\theta_i^1$ (filled black arrow), followed by Newton--Raphson correction (dotted path) toward the next admissible configuration (red arrow).
(C) Equivalent representation of the local update process viewed in the reduced plane spanned by $\Delta\theta_i^0$ and $\Delta\theta_i^1$, highlighting the geometric relationship between unconstrained updates, constraint projection, and convergence.}
\end{figure}

To clarify how these feasible folding trajectories are obtained, we first summarize the underlying principle of the constraint-matrix–based solver introduced by \textcite{hu2020}, upon which our framework directly builds.  
As schematically illustrated in Figure~\ref{fig:newton_raphson}A, the $n$-dimensional hinge-variable space contains a lower-dimensional constraint-satisfying region defined by loop-closure conditions.  
This region, referred to here as the \emph{reachable space} and depicted in ivory, represents all admissible rigid-folding configurations.  
Although this reachable space cannot be explicitly identified or enumerated in high-dimensional settings, feasible folding trajectories can be explored by steering the configuration within it using prescribed neutral angles and joint stiffnesses.

Given a target neutral angle $\bar{\theta}$ (red star in Figure~\ref{fig:newton_raphson}A) and assigned crease stiffness values, an initial update direction $\Delta\theta_i^0$ is computed from the current configuration.  
This direction reflects the energetic tendency toward the target configuration but does not, in general, satisfy the loop-closure constraints.  
The solver therefore iteratively projects this update onto the reachable space, yielding a constraint-consistent increment that guides the configuration along a feasible folding trajectory.  
By varying the target neutral angle $\bar{\phi}$ (blue star) and stiffness distribution, the same framework can reach multiple distinct configurations that all satisfy the closure constraints, revealing the multiplicity of admissible folding paths.

The local update mechanism is illustrated in Figures~\ref{fig:newton_raphson}B and~\ref{fig:newton_raphson}C.  
At a converged configuration $\theta_i^*$, the reachable space is locally approximated by a linearized constraint plane characterized directly by the constraint matrix $A(\theta_i^*)$.  
An initial unconstrained update toward the target neutral angle, denoted $\Delta\theta_i^0$, is first computed based on energetic preference and is depicted as an open black arrow.  
Because this update generally violates loop-closure constraints, it is orthogonally projected onto the linearized constraint plane (solid black line), producing a constraint-consistent increment $\Delta\theta_i^1$, shown as a filled black triangular arrow, which satisfies
$A(\theta_i^*)\,\Delta\theta_i^1 = 0$.

Residual constraint violations are subsequently reduced through a Newton--Raphson iteration, visualized as the dotted trajectory converging toward the ideal reachable space.  
The final converged update, indicated by a red triangular arrow, defines the next admissible configuration $\theta_{i+1}^*$.  
This local projection-and-correction process is repeated iteratively to generate a continuous folding trajectory that remains fully consistent with loop-closure constraints.

Building on this principle, we explored feasible folding trajectories by systematically varying the prescribed neutral angles and joint stiffness distributions.  
Our open-source implementation accepts a rigid-foldable pattern together with user-defined neutral angles and stiffness values, automatically computes constraint-consistent folding angles, and provides direct visualization of the resulting configurations.  
Figure~\ref{fig:rfs_examples} presents representative visualization results across diverse rigid-foldable structures, highlighting the range of geometric complexity, kinematic challenges, and folding behaviors that can be handled within a unified framework.

Figure~\ref{fig:rfs_examples}A shows a triangular Resch pattern, a class of tessellated rigid origami known to deform between two distinct flat states \parencite{resch1970design, tachi2013designing, yu2023programming}.  
While these end states are well documented, the transitional folding behavior connecting them is far less intuitive, and the kinematics of large Resch tessellations remain relatively unexplored.  
Despite the presence of 552 creases, the proposed framework robustly constructed the constraint matrix and successfully computed feasible folding trajectories across a wide range of intermediate configurations.

By adjusting the neutral angles and stiffness values assigned to individual creases, the solver revealed multiple distinct folding paths leading to the same final flat configuration.  
In the upper row of Figure~\ref{fig:rfs_examples}A, all creases are activated uniformly, producing a rounded triangular shape with curvature developing along three principal directions before returning to a flat state.  
In contrast, the lower row illustrates a trajectory in which curvature develops predominantly along a single direction: the pattern rolls cylindrically around an axis connecting opposite edges of the hexagonal boundary before reaching the fully folded flat state.  
Because neutral angles and stiffness values are assigned directly using edge indices generated during tessellation, this process remains intuitive and readily automatable even for large-scale patterns.

Figure~\ref{fig:rfs_examples}B presents a square pop-up kirigami mechanism consisting of a central square facet connected to four identical legs, each composed of five creases \parencite{yoneda2022structure}.  
Based on geometric observation, estimated neutral angles were assigned to the creases with uniform stiffness across the structure.  
The first row of Figure~\ref{fig:rfs_examples}B shows an axis-symmetric deployed configuration obtained by assigning identical neutral angles to creases of the same type on each leg.

Starting from this axis-symmetric deployed state, we deliberately broke the symmetry to explore alternative feasible deformation modes.  
In the second row of Figure~\ref{fig:rfs_examples}B, the leftmost image corresponds to the symmetric reference configuration, which is treated as a new initial state.  
By guiding the solver toward the flat configuration as a new terminal state and biasing intermediate updates to selectively lower certain legs while elevating others, multiple asymmetric configurations were obtained.

In these visualizations, the semi-transparent ivory geometry denotes the symmetric reference configuration, while the colored geometries (red and blue) represent two distinct asymmetric modes.  
The black arrow indicates the normal vector of the central square facet in the symmetric state, aligned with the vertical $z$-axis, whereas the colored arrows indicate the corresponding normals in the asymmetric configurations.  
In one mode, the central plate tilts forward relative to the viewing direction, while in the other it tilts laterally to the right.  
Such symmetry-breaking deployable motions are especially difficult to analyze analytically, as they involve coupled rotations and translations that invalidate simple trigonometric descriptions.  
In contrast, the proposed algorithm successfully captured the full loop-closure constraints and identified feasible folding configurations within the nonlinear reachable space while maintaining low residual errors, thereby preserving structural integrity throughout the deformation.

Figure~\ref{fig:rfs_examples}C demonstrates a stacked Miura origami tessellation, representing a hinging-type multi-sheet structure defined using the extended data schema \parencite{schenk2013}.  
For such configurations, manually identifying all independent closed loops is not only tedious but also highly error-prone; omission of even a single loop can distort the entire constraint relationship and invalidate feasible-angle computation.  
Moreover, traversing constraints across multiple sheets requires explicit treatment of translational components in loop-closure conditions, similar to kirigami structures, and the introduction of additional side hinges in three-dimensional layouts further complicates sign conventions.

Using the proposed data schema, each sheet-level pattern is generated independently, and inter-sheet hinging connections are specified directly between facet pairs.  
The framework maps these relationships into a facet–hinge graph and automatically extracts a minimal set of independent closure loops by computing the shortest cycles required for complete constraint enforcement.  
Additional hinges arising between sheets are detected automatically, and their sign conventions and joint orientations are consistently assigned based on vertex-index ordering.  
As shown in Figure~\ref{fig:rfs_examples}C, the framework robustly constructed constraint matrices for a structure with 144 creases and successfully identified feasible configurations despite the system possessing only a single degree of freedom and an extremely narrow reachable space.

Figure~\ref{fig:rfs_examples}D presents a TMP tessellation, corresponding to soldering-type connections with zero separation distance \parencite{yasuda2015reentrant}.  
As in stacked Miura origami, this structure involves multiple sheets connected across layers, and therefore shares similar challenges in identifying closed loops and constructing constraint matrices that correctly couple rotational and translational motion.  
In this case, soldered facet pairs are treated as a single rigid body, and no additional hinges are introduced.  
The facet–hinge graph correctly collapses these facets into unified nodes, yielding an exact yet reduced kinematic representation.  
Because mountain–valley sign conventions are inherited directly from the data schema and applied consistently on a per-sheet basis, feasible folding motions can be predicted purely from the sign pattern of the hinge variables.

Figure~\ref{fig:rfs_examples}E illustrates thick-panel implementations of the Miura and Yoshimura patterns generated using a hinge-shifting technique.  
Here, pairs of facets are specified as soldered with a finite offset distance, representing thick panels connected by shifted hinges.  
When such soldered face pairs are parallel and separated by a finite offset, the framework automatically generates the necessary side panels and corresponding kinematic constraints from the geometric definition, without requiring explicit specification of auxiliary faces.  
Both thick Miura and thick Yoshimura patterns yielded valid folding trajectories, demonstrating that the proposed formulation naturally accommodates thickness effects while preserving rigid-foldability.

\begin{figure}[p]  
\centering
\includegraphics[width=\linewidth]{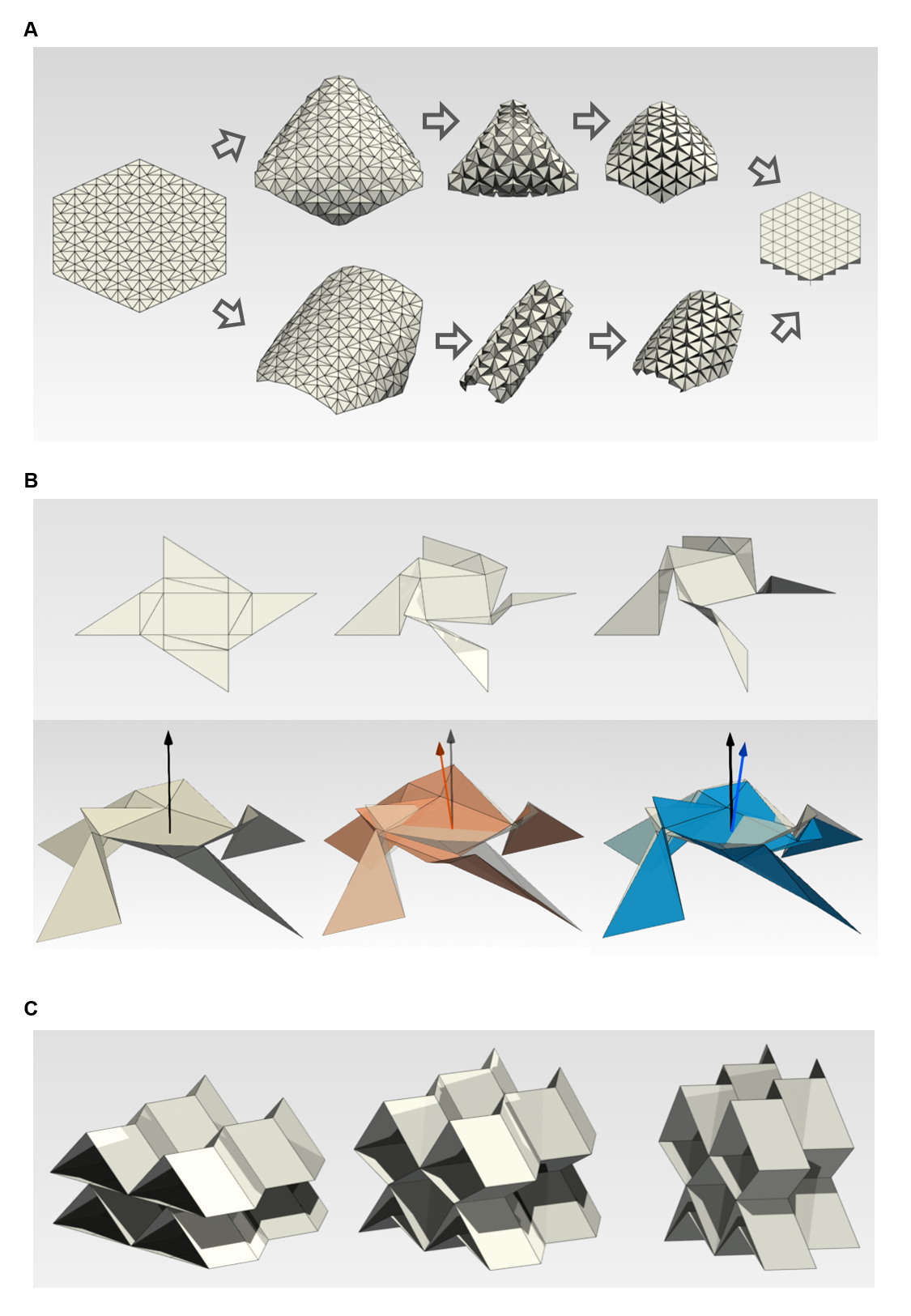}
\caption{Folding simulations across representative rigid-foldable structures. 
(A) A 4-orbit triangular Resch pattern exhibiting distinct folding trajectories. 
(B) A square pop-up kirigami mechanism. Top: axis-symmetric deploy motion. Bottom: two asymmetric modes (red and blue) compared with the symmetric configuration (ivory). 
(C) A stacked Miura origami tessellation.}
\label{fig:rfs_examples}
\end{figure}

\begin{figure}[p]\ContinuedFloat
\centering
\includegraphics[width=\linewidth]{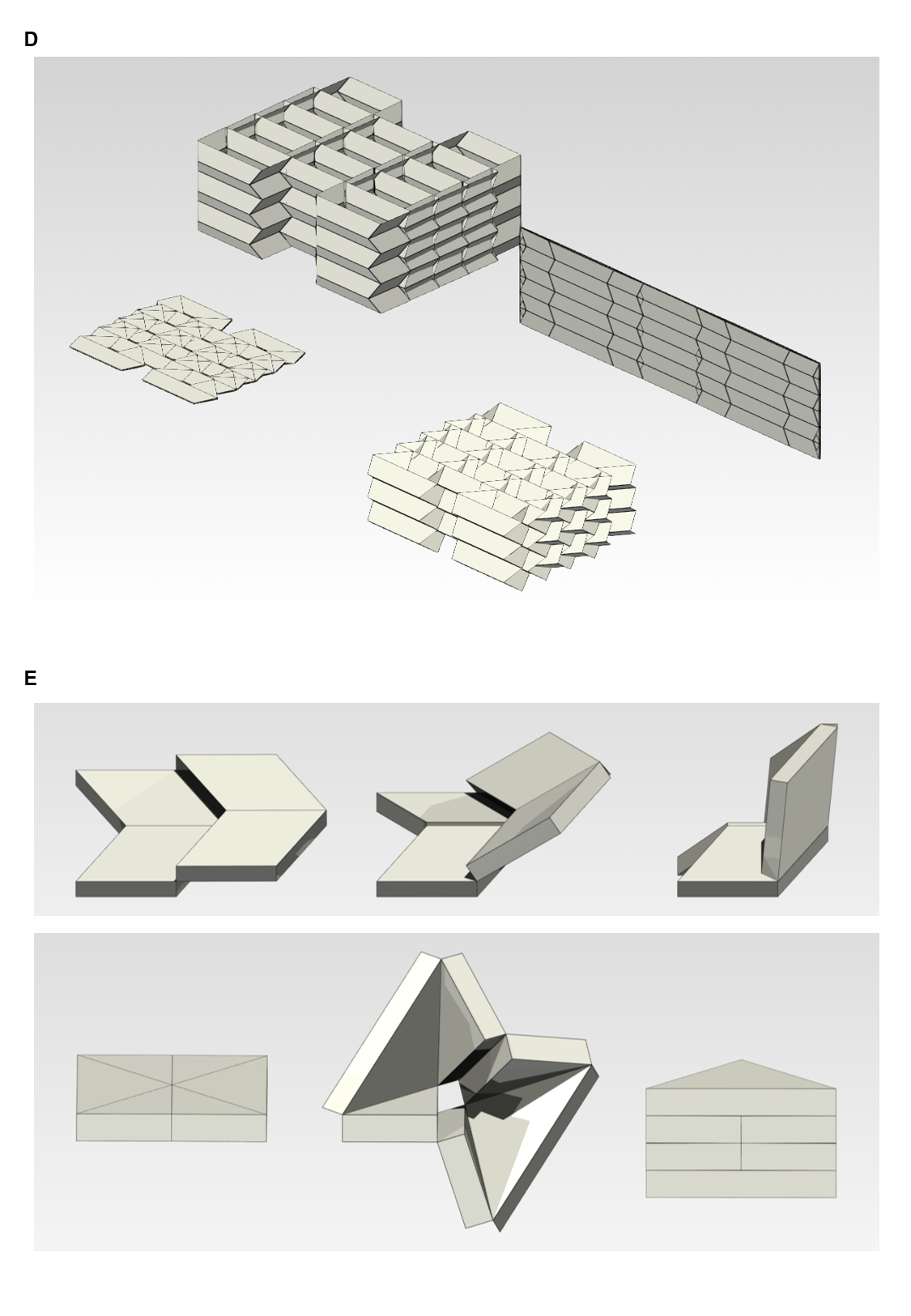}
\caption{Continuation of Figure~\ref{fig:rfs_examples}. 
(D) A TMP tessellation with soldered facets. 
(E) Thick-panel implementations of the Miura and Yoshimura patterns generated via hinge shifting.}
\end{figure}

These results demonstrate that the proposed framework can robustly construct constraint matrices and recover feasible folding trajectories across a wide spectrum of rigid-foldable structures, including large-scale tessellations, kirigami with translational motion, multi-sheet assemblies, and thick-panel implementations.  
To understand how such generality and robustness are achieved with minimal user intervention, we next discuss the underlying data schema and modeling philosophy that distinguish the proposed approach from existing representations.
While the widely adopted FOLD format \parencite{demaine2016fold} provides an efficient and portable representation of origami geometry, it requires explicit per-edge type assignments (e.g., mountain, valley, or unknown) and pairwise layer-order specifications between faces.  
Similarly, robotic modeling frameworks such as the URDF (Unified Robot Description Format) \parencite{urdf} offer standardized descriptions of articulated mechanisms but are inherently limited to tree-structured kinematic chains.  
To represent mechanisms with closed loops, extensions such as URDF+ \parencite{urdfplus} introduce explicit loop-joint and coupling declarations, together with parsers that reconstruct the corresponding closure constraints.  
However, these approaches still depend on user-defined identification of which joints complete the kinematic cycles, rather than inferring those loops and constraints directly from the geometry itself.  
Taken together, these geometry- and mechanism-level representations rely heavily on manual annotations---whether edge and layer assignments in FOLD or loop-joint declarations in URDF+.  

In contrast, our schema eliminates this manual step by defining mountain folds as positive and valley folds as negative hinge rotations, while automatically extracting loop structures from the given geometric information.  
This enables global consistency of folding motion to emerge naturally without additional user input, substantially reducing modeling effort while preserving full kinematic fidelity.  
Moreover, the schema remains compatible with common visualization standards such as the VTK format, allowing seamless transfer between analysis and rendering environments.

While the proposed data schema provides a compact and expressive description of geometry and connectivity, a principled kinematic formulation is still required to translate this information into enforceable motion constraints.  
To this end, we employ screw theory to derive a unified, velocity-level representation of rigid-folding kinematics directly from the geometric primitives encoded in the schema.
Specifically, the following formulation expresses rigid-folding kinematics in a compact and coordinate-free manner.
Because rigid-body transformations and their adjoint mappings belong to the same Lie algebraic structure, configuration updates, velocity propagation, and constraint construction can all be expressed within a unified operator framework.  
Once a configuration $\theta$ is specified, the corresponding Jacobian—and therefore the rows of the Pfaffian constraint matrix—are obtained directly through adjoint transformations.
Each row has a clear geometric interpretation as a local rigid-body constraint, enhancing both the intuitiveness and transparency of the formulation. Unlike conventional trigonometric or DH approaches that rely on manually defined local coordinates, this method operates directly on geometric primitives in initial configuration encoded in a minimal data schema.

\section{\label{sec:conclusion} Conclusion}
This study presented a unified kinematic framework for RFSs that systematically integrates geometric data, graph analogy, and screw-based constraint formulation.  
Here we explain three unique aspects of our unified framework for rigid foldable origami simulations.
First, the proposed data schema directly encodes facets, hinges, and their inter-sheet connections in a compact and extensible form.  
This representation eliminates the need for explicit mountain–valley or pairwise face-order assignments, since hinge rotations are defined consistently by sign conventions and sheet-wise orientation.  
Such structure allows straightforward extension to multi-layer and thick-panel configurations without modifying the underlying formulation.

Second, the facet–hinge graph provides an explicit representation of kinematic connectivity.
By extracting a minimum cycle basis, the method identifies a complete yet non-redundant set of closure loops that capture all independent kinematic constraints.  
This graph-theoretic perspective ensures scalability to large, tessellated foldable systems where explicit constraint definition would otherwise be intractable.

Third, screw alignment along each loop enables the automatic construction of the configuration-dependent constraint matrix.  
By integrating rotational and translational motion within a single representation, the framework provides a compact, coordinate-free means to compute the Pfaffian constraint matrix and analyze feasible folding directions at any configuration.  
Together, these components form a unified and geometrically intuitive approach for modeling the kinematics of rigid-foldable structures, offering both conceptual clarity and computational scalability.

Despite these advantages, several limitations remain.  
The current formulation assumes perfectly rigid panels and idealized hinges, neglecting elastic deformations that may arise in physical implementations.  
In addition, for mechanisms with a single degree of freedom but many interconnected hinges, the feasible motion subspace can be extremely narrow, making configuration search computationally demanding.  
While the constraint matrix characterizes instantaneous admissible motions, it does not by itself guarantee global reachability across the entire configuration space.  
Nevertheless, the automated construction of $A(\theta)$ significantly simplifies this search process, serving as a powerful tool for exploring feasible folding motions and initializing more advanced numerical solvers.

Finally, while screw theory provides strong geometric and physical intuition, it can be computationally demanding for large-scale systems.  
This observation motivates future work on developing more numerically efficient formulations that retain the same geometric foundation while improving scalability for high-dimensional simulations.

Beyond kinematic analysis, the proposed formulation also opens clear directions for future dynamic studies.  
Because the constraint matrix is expressed explicitly at the velocity level, it naturally provides a foundation for dynamic simulation, control, and optimization frameworks that incorporate inertia, actuation, and hinge torsion effects.

\section*{Conflicts of Interest} 
The authors declare no conflict of interest.

\section*{Author Contributions}
\noindent\textbf{Dongwook Kwak}: Conceptualization, Methodology, Software, Writing-original draft, Visualization. \textbf{Geonhee Cho}: Methodology, Software, Visualization, Writing-review \& editing. \textbf{Jiook Chung}: Software, Visualization, Writing-review \& editing.
\textbf{Jinkyu Yang}: Supervision, Funding Acquisition, Writing-review \& editing
\section*{Funding}
We acknowledge the support from Air Force Office of Scientific Research (FA2386-24-1-4051), SNU-IAMD, SNU-IOER, and National Research Foundation grants funded by the Korean government (2023R1A2C2003705 and 2022H1D3A2A03096579).

\section*{Data Availability}
All codes, example files, and usage demonstrations developed in this study are publicly available as open-source resources on our GitHub repository:
\url{https://github.com/dongwookkwak/Kinematic_Simulation_of_Rigid_Foldable_Structures}.
The repository provides complete implementations of the proposed pipeline, along with reproducible examples for constructing and analyzing rigid-foldable structures' kinematics.

\section*{Acknowledgments}
We thank Professor Tomohiro Tachi (University of Tokyo) and Professor  Glaucio Paulino’s team (Princeton University) for their valuable discussions.

\printbibliography

@article{zhang2023,
author = {Zhang, Yuehao and Li, Ming and Chen, Yan and Peng, Rui and Zhang, Xiao},
year = {2023},
month = {01},
pages = {105233},
title = {Thick-panel origami-based parabolic cylindrical antenna},
volume = {182},
journal = {Mechanism and Machine Theory},
doi = {10.1016/j.mechmachtheory.2023.105233}
}

@article{sun2024,
author = {Sun, Huize and Zhao, Chong and Wang, Ke and Zhao, Haifeng and Hongye, Ma and Zhang, Lu and Xue, Jing and Zhang, Lixian},
year = {2024},
month = {01},
pages = {105471},
title = {Shape editing of kirigami-inspired thick-panel deployable structure},
volume = {191},
journal = {Mechanism and Machine Theory},
doi = {10.1016/j.mechmachtheory.2023.105471}
}

@article{lerner2024,
author = {Lerner, Elisha and Chen, Zhe and Zhao, Jianguo},
year = {2024},
month = {10},
pages = {},
title = {Reconfigurable origami with variable stiffness joints for adaptive robotic locomotion and grasping},
volume = {382},
journal = {Philosophical Transactions A},
doi = {10.1098/rsta.2024.0017}
}

@article{chen2022,
author = {Chen, Zhe and Tighe, Brandon and Zhao, Jianguo},
year = {2022},
month = {08},
pages = {1-10},
title = {Origami-Inspired Modules Enable A Reconfigurable Robot with Programmable Shapes and Motions},
volume = {27},
journal = {IEEE/ASME Transactions on Mechatronics},
doi = {10.1109/TMECH.2022.3175145}
}

@article{zhu2024,
author = {Zhu, Yi and Filipov, Evgueni},
year = {2024},
month = {03},
pages = {},
title = {Large-scale modular and uniformly thick origami-inspired adaptable and load-carrying structures},
volume = {15},
journal = {Nature Communications},
doi = {10.1038/s41467-024-46667-0}
}

@article{hu2020,
title = {Folding simulation of rigid origami with Lagrange multiplier method},
journal = {International Journal of Solids and Structures},
volume = {202},
pages = {552-561},
year = {2020},
issn = {0020-7683},
doi = {https://doi.org/10.1016/j.ijsolstr.2020.06.016},
url = {https://www.sciencedirect.com/science/article/pii/S0020768320302389},
author = {Yucai Hu and Haiyi Liang}
}

@article{resch1970design,
  author = {Resch, R. D. and Christiansen, H.},
  journal = {Proceedings of IASS Symposium on Folded Plates and Prismatic Structures},
  title = {The Design and Analysis of Kinematic Folded Plate Systems },
  year = 1970
}

@article{tachi2013designing,
  title={Designing freeform origami tessellations by generalizing Resch's patterns},
  author={Tachi, Tomohiro},
  journal={Journal of mechanical design},
  volume={135},
  number={11},
  pages={111006},
  year={2013},
  publisher={American Society of Mechanical Engineers}
}

@article{yu2023programming,
title = {Programming curvatures by unfolding of the triangular Resch pattern},
journal = {International Journal of Mechanical Sciences},
volume = {238},
pages = {107861},
year = {2023},
issn = {0020-7403},
doi = {https://doi.org/10.1016/j.ijmecsci.2022.107861},
url = {https://www.sciencedirect.com/science/article/pii/S0020740322007408},
author = {Ying Yu and Yan Chen and Glaucio Paulino}
}

@article{yoneda2022structure,
author = {Yoneda, Taiju and Wada, Hirofumi},
year = {2022},
pages = {L021004},
title = {Structure, Design, and Mechanics of a Pop-Up Origami with Cuts},
volume = {17},
journal = {Physical Review Applied},
number={2},
doi = {10.1103/PhysRevApplied.17.L021004},
publisher={APS}
}

@article{yasuda2015reentrant,
  title = {Reentrant Origami-Based Metamaterials with Negative Poisson's Ratio and Bistability},
  author = {Yasuda, H. and Yang, J.},
  journal = {Phys. Rev. Lett.},
  volume = {114},
  issue = {18},
  pages = {185502},
  numpages = {5},
  year = {2015},
  publisher = {American Physical Society},
  doi = {10.1103/PhysRevLett.114.185502},
  url = {https://link.aps.org/doi/10.1103/PhysRevLett.114.185502}
}

@inproceedings{demaine2016fold,
  author    = {Erik D. Demaine, Jason S. Ku and Robert J. Lang},
  title     = {A New File Standard to Represent Folded Structures},
  booktitle = {Abstracts from the 26th Fall Workshop on Computational Geometry},
  year      = {2016},
  address   = {Brandeis University, Waltham, MA}
}

@article{zhu2022,
author = {Zhu, Yi and Schenk, Mark and Filipov, Evgueni},
year = {2022},
month = {07},
pages = {},
title = {A Review On Origami Simulations: From Kinematics, to Mechanics, Towards Multi-Physics},
volume = {74},
journal = {Applied Mechanics Reviews},
doi = {10.1115/1.4055031}
}

@article{yasuda2013,
  title={Folding behaviour of Tachi--Miura polyhedron bellows},
  author={Yasuda, Hiromi and Yein, Thu and Tachi, Tomohiro and Miura, Koryo and Taya, Minoru},
  journal={Proceedings of the Royal Society A: Mathematical, Physical and Engineering Sciences},
  volume={469},
  number={2159},
  pages={20130351},
  year={2013},
  publisher={The Royal Society Publishing}
}

@article{schenk2013,
  title={Geometry of Miura-folded metamaterials},
  author={Schenk, Mark and Guest, Simon D},
  journal={Proceedings of the National Academy of Sciences},
  volume={110},
  number={9},
  pages={3276--3281},
  year={2013},
  publisher={National Academy of Sciences}
}

@article{
chen2016,
author = {Chen, Yan  and Feng, Huijuan  and Ma, Jiayao  and Peng, Rui  and You, Zhong },
title = {Symmetric waterbomb origami},
journal = {Proceedings of the Royal Society A: Mathematical, Physical and Engineering Sciences},
volume = {472},
number = {2190},
pages = {20150846},
year = {2016},
doi = {10.1098/rspa.2015.0846},
}

@article{
chen2015,
author = {Yan Chen  and Rui Peng  and Zhong You },
title = {Origami of thick panels},
journal = {Science},
volume = {349},
number = {6246},
pages = {396-400},
year = {2015},
doi = {10.1126/science.aab2870},
URL = {https://www.science.org/doi/abs/10.1126/science.aab2870},
eprint = {https://www.science.org/doi/pdf/10.1126/science.aab2870}}

@article{wang2024,
title = {Deployment dynamics of thick panel Miura-origami},
journal = {Aerospace Science and Technology},
volume = {144},
pages = {108795},
year = {2024},
issn = {1270-9638},
doi = {https://doi.org/10.1016/j.ast.2023.108795},
url = {https://www.sciencedirect.com/science/article/pii/S1270963823006910},
author = {Cheng Wang and Dawei Zhang and Junlan Li and Yingjie Li and Xiaofeng Zhang},
}

@article{yang2022,
title = {Design of Single Degree-of-Freedom Triangular Resch Patterns with Thick-panel Origami},
journal = {Mechanism and Machine Theory},
volume = {169},
pages = {104650},
year = {2022},
issn = {0094-114X},
doi = {https://doi.org/10.1016/j.mechmachtheory.2021.104650},
url = {https://www.sciencedirect.com/science/article/pii/S0094114X21003815},
author = {Fufu Yang and Miao Zhang and Jiayao Ma and Zhong You and Ying Yu and Yan Chen and Glaucio H. Paulino},
}

@article{tachi2009,
  title={Simulation of Rigid Origami},
  author={Tachi, Tomohiro},
  journal={Origami 4},
  pages={175},
  year={2009},
  publisher={CRC Press}
}

@article{suto2023,
author = {Suto, Kai and Noma, Yuta and Tanimichi, Kotaro and Narumi, Koya and Tachi, Tomohiro},
title = {Crane: An Integrated Computational Design Platform for Functional, Foldable, and Fabricable Origami Products},
year = {2023},
issue_date = {August 2023},
publisher = {Association for Computing Machinery},
address = {New York, NY, USA},
volume = {30},
number = {4},
issn = {1073-0516},
url = {https://doi.org/10.1145/3576856},
doi = {10.1145/3576856},
journal = {ACM Trans. Comput.-Hum. Interact.},
month = sep,
articleno = {52},
numpages = {29},

}

@article{liu2017,
    author = {Liu, K. and Paulino, G. H.},
    title = {Nonlinear mechanics of non-rigid origami: an efficient computational approach†},
    journal = {Proceedings of the Royal Society A: Mathematical, Physical and Engineering Sciences},
    volume = {473},
    number = {2206},
    pages = {20170348},
    year = {2017},
    month = {10},
    issn = {1364-5021},
    doi = {10.1098/rspa.2017.0348},
    url = {https://doi.org/10.1098/rspa.2017.0348},
    eprint = {https://royalsocietypublishing.org/rspa/article-pdf/doi/10.1098/rspa.2017.0348/365835/rspa.2017.0348.pdf},
}

@article{ghassaei2018,
  title={Fast, interactive origami simulation using GPU computation},
  author={Ghassaei, Amanda and Demaine, Erik D and Gershenfeld, Neil},
  journal={Origami},
  volume={7},
  pages={1151--1166},
  year={2018}
}

@book{
modern_robotics, 
author = {Lynch, Kevin M. and Park, Frank C.}, 
title = {Modern Robotics: Mechanics, Planning, and Control}, 
year = {2017}, 
isbn = {1107156300}, 
publisher = {Cambridge University Press}, 
address = {USA}, 
edition = {1st}}

@article{Horton1987,
  author  = {J. D. Horton},
  title   = {A polynomial-time algorithm to find the shortest cycle basis of a graph},
  journal = {SIAM Journal on Computing},
  volume  = {16},
  number  = {2},
  pages   = {358--366},
  year    = {1987},
  doi     = {10.1137/0216024}
}

@book{urdf,
author = {Quigley, Morgan and Gerkey, Brian and Smart, William D.},
title = {Programming Robots with ROS: A Practical Introduction to the Robot Operating System},
year = {2015},
isbn = {1449323898},
publisher = {O'Reilly Media, Inc.},
edition = {1st}
}

@INPROCEEDINGS{urdfplus,
  author={Chignoli, Matthew and Slotine, Jean-Jacques and Wensing, Patrick M. and Kim, Sangbae},
  booktitle={2024 IEEE-RAS 23rd International Conference on Humanoid Robots (Humanoids)}, 
  title={URDF+: An Enhanced URDF for Robots with Kinematic Loops}, 
  year={2024},
  volume={},
  number={},
  pages={197-204},
  doi={10.1109/Humanoids58906.2024.10769903}}

\renewcommand\theequation{\Alph{section}\arabic{equation}} 
\counterwithin*{equation}{section} 
\renewcommand\thefigure{\Alph{section}\arabic{figure}} 
\counterwithin*{figure}{section} 
\renewcommand\thetable{\Alph{section}\arabic{table}} 
\counterwithin*{table}{section} 
\begin{appendices}
\section*{Appendix}

\section{\label{appendix:flow} Computational Pipeline for Kinematic Constraint Generation}
The kinematic analysis of an arbitrary Rigid Foldable Structure (RFS) is automated through a computational pipeline that transforms a geometric definition into a complete velocity-level constraint matrix. This process ensures that all kinematic dependencies, including coupled rotation and translation, are systematically captured. The pipeline proceeds through four principal stages: graph construction from the data schema, minimal loop extraction, consistent screw assignment, and final matrix assembly using screw theory.

\subsection{\label{appendix:convention}Convention for Data Schema}

Each sheet designates a seed facet whose orientation (clockwise or counterclockwise) serves as the reference for the entire sheet.  
All facets within a sheet must be defined as valid, non-self-intersecting polygons, and their vertex indices must be ordered consistently in a single direction so that the sheet maintains a uniform orientation.  
Once the seed facet is defined, the orientations of all remaining facets are automatically aligned through shared pattern edges:  
for each pair of adjacent facets, the shared edge must appear in opposite vertex order within their respective vertex lists  
(e.g., $[\cdots, v_a, v_b, \cdots]$ in one facet and $[\cdots, v_b, v_a, \cdots]$ in its neighbor).  
This opposite ordering enables the algorithm to infer and propagate consistent facet orientations across the sheet based solely on vertex indices,  
ensuring that all hinge rotations follow a coherent mountain–valley convention without additional user input.

When multiple facets from one sheet are connected to another sheet, all such connections must share the same connection type $x \in \{h, s\}$.  
For example, if facet $f_{00}$ of sheet 0 and facet $f_{10}$ of sheet 1 are joined through a hinging-type connection ($x=h$),  
all remaining facet pairs between the same two sheets must also be of type $h$.  
Mixed connection types—such as $[f_{00}, f_{10}, h]$ and $[f_{01}, f_{11}, s]$ between the same sheet pair—are not permitted.  
In such cases, the sheet should instead be divided into separate sub-sheets with distinct connection types.

For hinging-type connections ($x = h$), an identical edge with coincident vertex coordinates must exist between the two facets.  
If no such shared edge is found, the connection is deemed invalid.  
If multiple coincident edges are detected between the same facet pair, the user may explicitly specify which edge index serves as the active hinge.  
That selected edge is treated as the valid rotational joint, while all other overlapping edges are regarded as non-hinging pattern edges.  
If the user does not specify an active hinge, multiple coincident hinging edges are automatically treated as a soldering-type connection, and an error is reported.  
In all cases, one facet pair must correspond to at most one hinging edge.

For cylindrical, spherical, or otherwise closed-topology patterns, the geometry may be divided into two or more sheets along non-creasing boundaries.  
These boundary edges can then be reconnected using hinging-type links, enabling the framework to assign consistent mountain–valley conventions within each sheet while maintaining global kinematic consistency.
However, mountain–valley labeling cannot be intuitively preserved across multiple sheets in three-dimensional space,  
because the distinction between “top” and “bottom” depends on the observer’s perspective and the local embedding of each sheet.  
As a result, the same physical rotation may appear as a mountain fold in one viewpoint and a valley fold in another.
To address this, the algorithm checks inter-sheet consistency and, whenever a discrepancy arises, reports the positive hinge-axis orientation associated with each user-specified connection record $[f_1, f_2, h]$.  
By explicitly printing the screw direction from facet $f_1$ toward $f_2$, the framework enables users to confirm how the sign of a hinge variable corresponds to the actual folding motion. 
This provides both intuitive interpretability and geometric traceability of inter-sheet connections.

Facet pairs sharing only a single vertex or forming self-intersecting boundaries are excluded as invalid configurations when they belong to the same sheet (intra-sheet case).  
However, if such contacts occur between different sheets (inter-sheet case), they are automatically interpreted as soldering-type connections.

Before graph construction, the framework automatically verifies the validity of the data schema by checking  
(i) consistency of connection types, (ii) alignment of facet orientations inherited from each seed facet, and (iii) the presence of closed-loop validity.

\subsection{\label{appendix:graph_construction}Facet-Hinge Graph Construction from Data Schema}
The pipeline commences with the RFS Information Data Schema, the minimal geometric input. This schema is a layer-wise extension of the conventional (V, E, F, C) representation, explicitly defining vertices ($v_{ij}$), pattern edges ($e_{ij}$), and facets ($f_{ij}$) on a per-sheet basis. Crucially, it also includes a list of sheet-wise connections ($c_{\alpha}$) that specify how different layers' facets interact. This input is algorithmically converted into a Facet-Hinge Graph. In this graph, each facet $f_{ij}$ is represented as a node.

Graph edges are created based on connectivity:
\begin{enumerate}
    \item \textbf{Intra-sheet connections:} The algorithm identifies shared pattern edges. Since each facet $f_{ij}$ is an ordered list of vertex indices, and its orientation (e.g., CW or CCW) is defined consistently from a seed facet, a shared boundary between two adjacent facets is algorithmically identified by finding an oppositely ordered vertex pair (e.g., $(v_a, v_b)$ in one facet's list and $(v_b, v_a)$ in the other's). This shared edge becomes a graph edge. The 3D vertex coordinates $v_a, v_b \in \mathbb{R}^3$ directly define the initial screw parameters for this hinge: the rotation axis is computed as $\omega = (v_b - v_a) / \|v_b - v_a\|$, and a point on the axis is $q = v_a$ (or $v_b$).
    
    \item \textbf{Inter-sheet connections:} The connection list $c_{\alpha}$ specifies two facet indices and a type $x \in \{h, s\}$. If the type is $x=s$ (soldering), no screw is required, as the corresponding facet nodes are simply merged into a single node. If the type is $x=h$ (hinging), the algorithm performs a geometric comparison of the 3D coordinates of the vertices for both facets to identify shared boundary edges. A kinematic ambiguity arises if the facets share two or more non-collinear edges; hinging along multiple distinct axes would lock the relative motion, effectively creating a soldered connection. Therefore, if the geometric comparison identifies multiple potential hinge edges, the algorithm make user select only one of them to define the single inter-sheet screw axis. A new graph edge, representing this single revolute joint, is then created between the facet nodes.\end{enumerate}
    
This automated conversion translates the geometric data into a topological representation of kinematic connectivity, with all necessary screw parameters derived directly from the vertex coordinates.

\subsection{\label{appendix:screw_assign}Global Screw Assignment}
Once the independent loops are identified via the MCB, the set of hinges (graph edges) that participate in these basis loops are identified as the kinematically active screws. Hinges not belonging to any loop in the MCB are kinematically irrelevant and excluded from the set of $n$ joint variables, $\theta$. Each of these $n$ active screws must be assigned a consistent screw axis $\xi_i$ to represent its rotational motion. This step is critical for avoiding sign conflicts in the global constraint equations.

This assignment process relies on a fundamental precondition: the RFS, and all its constituent loops, must be \textbf{orientable}. This ensures that a globally consistent `front' and `back' (or `mountain'/`valley') can be defined without encountering a topological contradiction (e.g., a Möbius strip). Given this precondition, the assignment is a two-stage process:

\begin{enumerate}
    \item \textbf{Base Orientation (BO) Establishment:} A globally consistent relative orientation is algorithmically established. This is possible because the graph formed by the $L$ minimal cycles (where cycles are nodes and shared hinges are edges) is a connected graph. Therefore, a single seed facet in one minimal cycle is chosen to fix its traversal direction. This assignment propagates deterministically through the entire connected graph: once a hinge's orientation ($\xi_i$) is fixed, it forces the traversal direction of any other cycle in the MCB that contains it. This process repeats until all $n$ active screws possess an internally consistent relative orientation

    \item \textbf{Sheet Polarity Application:} The established BO is reconciled with the user-defined sheet polarity. The data schema allows a seed facet per sheet, which defines the user's desired `positive' rotation (e.g., mountain fold). The algorithm applies this per-sheet definition as a sign correction to the BO, subject to a validity check:
    \begin{itemize}
        \item \textbf{Valid Sheet:} A sheet is valid if all its constituent hinges are `in-phase' (i.e., must move in the same relative direction) according to the globally established BO. The user's seed can then validly assign a single polarity (e.g., `all mountain' or `all valley') to this entire group.
        \item \textbf{Invalid Sheet:} A sheet is invalid if it attempts to group hinges that the BO has already determined must move in `opposite-phase'. Such a definition is topologically contradictory, as it commands kinematically opposed elements to move as one.
    \end{itemize}
\end{enumerate}

The framework assumes all user-defined sheets are valid. The final set of $n$ screw axes $\{\xi_i\}$ thus reflects both the global kinematic consistency (from the BO) and the desired local folding polarity (from the valid sheet seeds).

\section{\label{appendix:ScrewTheory}Screw Theory for Rigid Body Motions}
This section introduces the mathematical foundation for describing rigid body motions and spatial kinematics using Lie group and Lie algebra formulations. Based on the Chasles–Mozzi theorem, we demonstrate that any rigid motion can be expressed as a screw motion, compactly encoded by a 6-dimensional screw axis. Using exponential coordinates, we provide closed-form expressions for transformations, which naturally lead to the Product of Exponentials (PoE) formulation for forward kinematics. We then introduce spatial velocities (twists) and adjoint mappings that enable frame transformations of motion representations. This framework also leads to the definition of the space Jacobian, which maps joint velocities to the space twist.  
Together, these tools form a coherent foundation for analyzing the rigid body kinematics of any rigid and foldable structures. The notations and formulations used in the following sections are based on the framework presented in \textcite{modern_robotics}, which also provides further details on the theoretical foundations of this approach.

\subsection{\label{sec:screw_motions} Exponential Coordinate Representations}
\begin{figure}[!ht]
\centering
\includegraphics[width=1\textwidth]{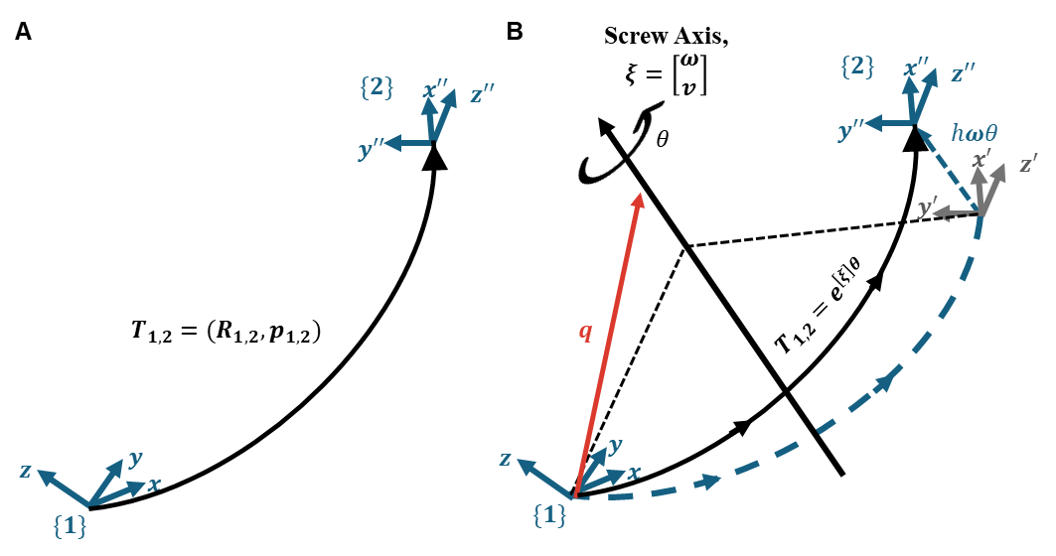}
\caption{Rigid body motion (A) Rigid body motion represented by transformation from frame $\{1\}$ to frame $\{2\}$ (B) Rigid body motion represented as a screw motion from frame $\{1\}$ to frame $\{2\}$}
\label{fig:RigidBodyMotion}
\end{figure}

Rigid body motion in three-dimensional space is commonly represented as an element of the Special Euclidean group $SE(3)$, which combines rotation and translation in a single homogeneous transformation matrix. Consider two coordinate frames $\{1\}$ and $\{2\}$. As illustrated in Figure~\ref{fig:RigidBodyMotion}A, the transformation from frame $\{1\}$ to frame $\{2\}$ can be expressed as:
\begin{equation}
T_{1,2} = \begin{bmatrix} R_{1,2} & p_{1,2} \\ 0 & 1 \end{bmatrix} \in SE(3),
\end{equation}
where $R_{1,2} \in SO(3)$ is a $3\times3$ rotation matrix belonging to the Special Orthogonal group, and $p_{1,2} \in \mathbb{R}^3$ is a translation vector.

For convenience, the transformation $T_{1,2}$ is often denoted as the pair $(R_{1,2}, p_{1,2})$. The subscripts ${1,2}$ indicate that this transformation maps coordinates from frame $\{1\}$ (the source frame) to frame $\{2\}$ (the target frame).

According to the Chasles–Mozzi theorem, any rigid body motion can be represented as a screw motion, illustrated in Figure~\ref{fig:RigidBodyMotion} (B). The screw motion consists of a rotation about a fixed axis and a translation along that axis. The amount of translation per unit rotation is called the pitch, denoted by $h$, which is equal to the ratio of linear speed to angular speed. In the case of a pure rotation (e.g., a revolute joint), the pitch $h$ becomes zero.
A screw motion is characterized by a \textit{screw axis} $\xi \in \mathbb{R}^6$, a 6-dimensional vector that combines rotational and translational motion:
\begin{equation}
\xi = \begin{bmatrix} \omega \\ v \end{bmatrix} \in \mathbb{R}^6,
\end{equation}
where $\omega \in \mathbb{R}^3$ is a vector representing the axis of rotation, and $v \in \mathbb{R}^3$ is computed as
\begin{equation}
v = -\omega \times q + h\omega,
\end{equation}
with $q$ being any point on the screw axis in the reference frame and $h$ denoting the pitch.

The 4-by-4 matrix representation of the screw $\xi$ is an element of the Lie algebra $se(3)$:
\begin{equation}
[\xi] = \begin{bmatrix} [\omega] & v \\ 0 & 0 \end{bmatrix} \in se(3),
\end{equation}
where $[\omega] \in so(3)$ is the skew-symmetric matrix corresponding to $\omega = (\omega_1, \omega_2, \omega_3)$:
\begin{equation}
[\omega] = \begin{bmatrix} 0 & -\omega_3 & \omega_2 \\ \omega_3 & 0 & -\omega_1 \\ -\omega_2 & \omega_1 & 0 \end{bmatrix}.
\end{equation}
Depending on whether the rotational component $\omega$ is nonzero, a screw motion falls into one of two cases.  
When $\|\omega\| = 1$, corresponding to a screw motion with rotation, the rigid-body transformation for a motion magnitude $\theta$ can be computed using the matrix exponential:
$$
T(\theta) = e^{[\xi]\theta} = 
\begin{bmatrix}
e^{[\omega]\theta} & G(\theta)v \\
0 & 1
\end{bmatrix}.
$$
The rotation matrix $e^{[\omega]\theta}$ is given by Rodrigues’ formula:
$$
e^{[\omega]\theta} = I_3 + \sin\theta [\omega] + (1 - \cos\theta)[\omega]^2.
$$
The matrix $G(\theta)$ couples the translational part of the motion and is defined as:
$$
G(\theta) = I_3\theta + (1 - \cos\theta)[\omega] + (\theta - \sin\theta)[\omega]^2,
$$
where $I_k$ denotes the $k \times k$ identity matrix.
In contrast, when $\|\omega\| = 0$, the screw represents a pure translation. In this case, $\|v\| = 1$, where $v$ is a unit vector that describes the translation direction. The transformation can be further reduced to:
$$
T(\theta) = 
\begin{bmatrix}
I_3 & v\theta \\
0 & 1
\end{bmatrix}.
$$

In this context, $SE(3)$ and $SO(3)$ are Lie groups representing smooth manifolds of rigid body transformations—$SE(3)$ for combined rotation and translation, and $SO(3)$ for pure rotation. Their associated Lie algebras, $se(3)$ and $so(3)$, correspond to instantaneous rigid body motions. The matrices $[\omega] \in so(3)$ and $[\xi] \in se(3)$ lie in these Lie algebras, and the exponential map provides the bridge from these algebraic structures to the finite rigid body motions described by the Lie groups.

\subsection{\label{sec:PoE} Forward Kinematics and Product of Exponentials}
\begin{figure}[hbt]
\centering
\includegraphics[width=0.5\linewidth]{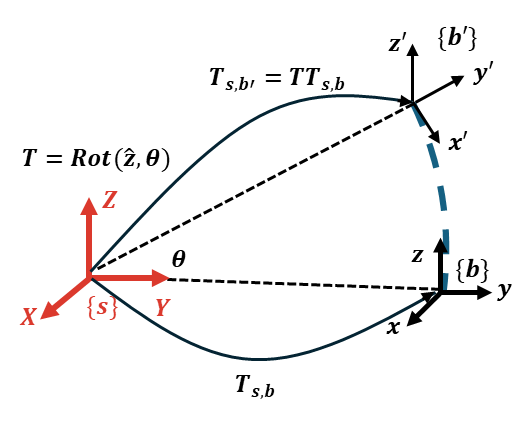}
\caption{Pre-multiplying the transformation matrix operates in the fixed frame}
\label{fig:pre_multiply}
\end{figure}

In Figure~\ref{fig:pre_multiply}, let $\{s\}$ denote a fixed space frame and $\{b\}$ a moving body frame. The configuration of the body frame relative to the space frame is represented by the transformation matrix $T_{s,b} \in SE(3)$. When a transformation matrix $T$ is applied via pre-multiplication, it is interpreted as a transformation with respect to the fixed space frame. For example, in the figure, if $T = \mathrm{Rot}(z, \theta)$ represents a rotation about the $z$-axis, then pre-multiplying $T$ to $T_{s,b}$ yields $T_{s,b'} = T\, T_{s,b}$, which corresponds to rotating the body frame $\{b\}$ about the global $Z$-axis of the space frame.
\begin{figure}[hbt]
\centering
\includegraphics[width=1\textwidth]{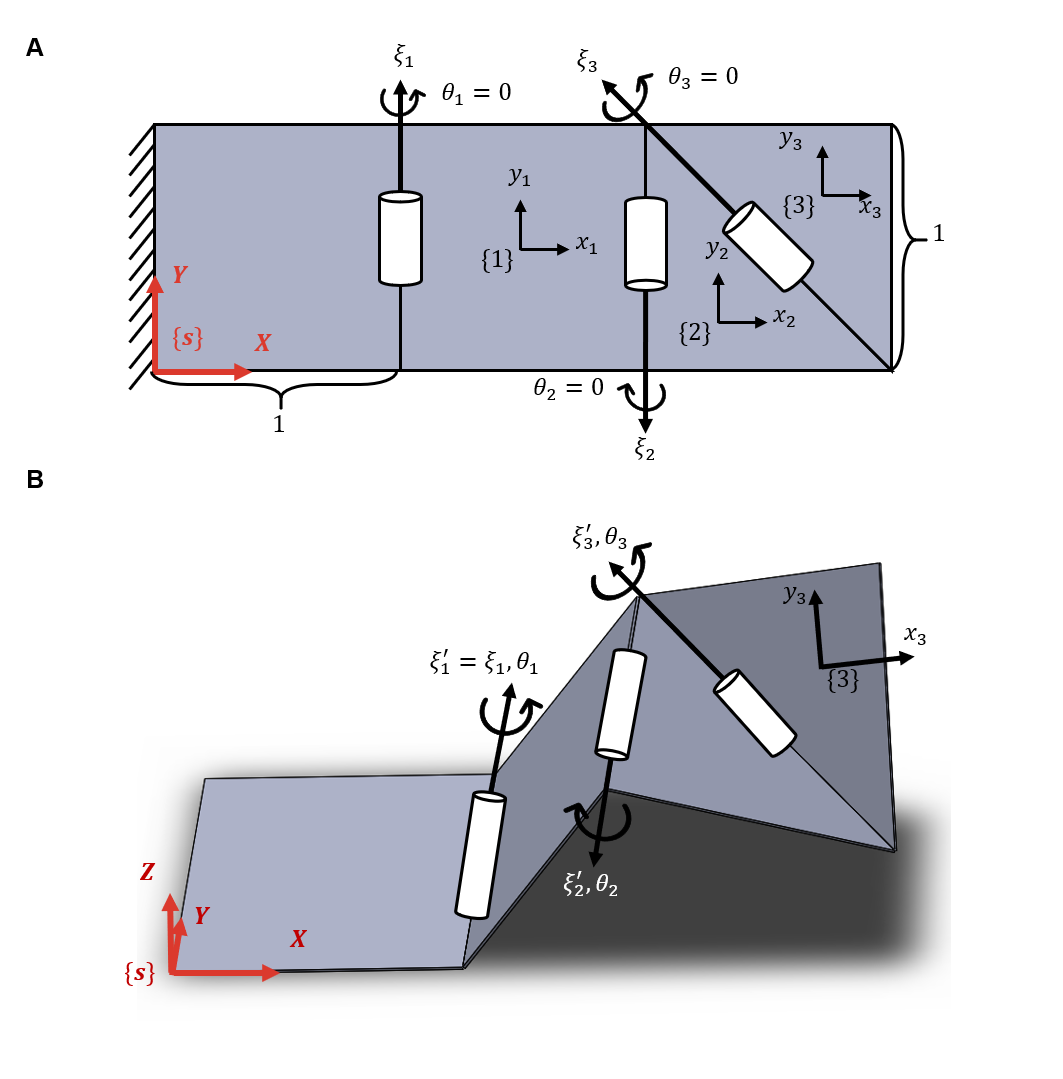}
\caption{Example of an open-chain origami (A) the home position (B) after crease rotation}
\label{fig:open_chain}
\end{figure}
Now, Consider the open-chain rigid origami structure shown in Figure~\ref{fig:open_chain}, which consists of four facets connected by three joints. The leftmost square facet is fixed to the ground and serves as the space frame; thus, it is not counted as part of the kinematic chain. The remaining three facets are treated as rigid links.

The configuration in which all crease (or joint) angles are zero, i.e., $\theta = (0, 0, 0)$, is referred to as the \textit{home position}. This typically corresponds to the initial unfolded pattern of a rigid-foldable structure, which is assumed to be known a priori. The home position provides a convenient and systematic way to initialize the model, as the screw axes and local facet frames can be derived directly from geometric information—namely, vertex coordinates and the connectivity among edges and facets. Since this geometric representation encodes the entire structural topology, the home configuration serves as a general and extensible foundation for automating screw parameter computation and coordinate frame assignment.

In this configuration, the screw axis definitions become straightforward. Since all creases are modeled as revolute joints, the screw pitch is $h = 0$ for all cases. The screw axis information defined in the space frame at the home position—hereafter referred to as the \textit{home position space screw axes}—is summarized in Table~\ref{tab:home_position_screw}.

\begin{table}[ht]
\caption{Example of screw axis assignment in home position\label{tab:home_position_screw}}
\begin{threeparttable}
\begin{tabular*}{\columnwidth}{@{\extracolsep\fill}cccc@{\extracolsep\fill}}
\toprule
Joint $i$ & $\omega_i$ & $q_i$ & $v_i = -\omega_i \times q_i$ \\
\midrule
1 & $(0, 1, 0)$ & $(1, 0, 0)$ & $(0, 0, 1)$ \\
2 & $(0, -1, 0)$ & $(2, 0, 0)$ & $(0, 0, -2)$ \\
3 & $(-1/\sqrt{2}, 1/\sqrt{2}, 0)$ & $(3, 0, 0)$ & $(0, 0, 3/\sqrt{2})$ \\
\bottomrule
\end{tabular*}
\end{threeparttable}
\end{table}

Each facet may be assigned an arbitrary local coordinate frame, as long as transformations are properly defined. Let $M_{s,3}$ denote the transformation matrix of the third facet in the home position. These matrices are referred to as the \textit{home position transformation matrices}, denoted by $M_{s,i}$.

Now consider the case where the creases are rotated by an amount $(\theta_1, \theta_2, \theta_3)$, resulting in the configuration shown in Figure~\ref{fig:open_chain}B. After the crease rotations, the new transformation matrix of the third facet, $T_{s,3}$, is given by:
\begin{equation}
    T_{s,3} = e^{[\xi_1]\theta_1}e^{[\xi_2]\theta_2}e^{[\xi_3]\theta_3}M_{s,3}.
\end{equation}

This expression follows from the property of pre-multiplying transformation matrices. It is known as the \textit{Product of Exponentials} (PoE) formula and provides a straightforward method for computing forward kinematics.

More generally, for an open chain with $n$ joints (and $n$ corresponding links), the forward kinematics using the PoE formulation becomes:
\begin{equation}
    T_{s,n} = e^{[\xi_1]\theta_1}e^{[\xi_2]\theta_2} \cdots e^{[\xi_n]\theta_n}M_{s,n},
    \label{eq:PoE}
\end{equation}
where the lower index of each screw axis indicates proximity to the base or space frame, and the higher index indicates proximity to the end-effector or terminal linkage.

Although the Denavit–Hartenberg (DH) convention can also be used to describe the forward kinematics of an open chain with $n$ joints using a minimal set of $4n$ parameters, it requires strict rules for attaching frames to each link. In contrast, the PoE formulation offers greater flexibility in frame assignment and is therefore preferred in this work.

\subsection{\label{sec:twist_jacobian}Space Twist and Jacobian}
For an arbitrary moving body frame observed from the space frame, the transformation matrix is denoted as $T_{s,b}(t) \in SE(3)$. The spatial velocity of the body—also referred to as the \textit{space twist}—can be expressed as:
\begin{equation}
    [\mathcal{V}_s] = \begin{bmatrix}
        [\omega_s] & v_s \\ 0 & 0
    \end{bmatrix} = \dot{T}_{s,b} T_{s,b}^{-1} \in se(3),
    \label{eq:space_twist}
\end{equation}
where $\mathcal{V}_s = \begin{bmatrix} \omega_s \\ v_s \end{bmatrix} \in \mathbb{R}^6$ is the twist vector expressed in the space frame $\{s\}$.  

The bracket operator $[\cdot]$ maps a 3-vector to a $3 \times 3$ skew-symmetric matrix in $so(3)$, and a 6-vector to a $4 \times 4$ matrix in $se(3)$. That is, for a twist $\mathcal{V}_s \in \mathbb{R}^6$, $[\mathcal{V}_s] \in se(3)$ is its matrix representation.

Note that a twist encodes the instantaneous motion of a rigid body and lies in the Lie algebra $se(3)$. A unit screw axis can be viewed as a normalized twist direction associated with a particular rigid body motion.

Physically, $\omega_s$ represents the angular velocity of the body frame with respect to the space frame, and $v_s$ is the linear velocity of the point on the body that instantaneously coincides with the origin of $\{s\}$.

The \textit{adjoint mapping} is an operator that enables the transformation of twists (or screw axes) across different reference frames. It serves two primary purposes: (1) expressing the same twist in a different reference frame, and (2) transforming a twist that has been displaced by a sequence of rigid body motions.

For an arbitrary homogeneous transformation matrix $T_{a,b} = (R_{a,b}, p_{a,b})$, the adjoint operator $\mathrm{Ad}_{T_{a,b}} : \mathbb{R}^6 \rightarrow \mathbb{R}^6$ is defined as:
\begin{equation}
    \mathcal{V}_a = \mathrm{Ad}_{T_{a,b}}(\mathcal{V}_b) = [\mathrm{Ad}_{T_{a,b}}] \mathcal{V}_b ,
    \label{eq:Adjoint_start}
\end{equation}
where
\begin{equation}
    [\mathrm{Ad}_{T_{a,b}}] = \begin{bmatrix}
        R_{a,b} & 0 \\
        [p_{a,b}]R_{a,b} & R_{a,b}
    \end{bmatrix} \in \mathbb{R}^{6 \times 6}.
\end{equation}
For twists represented in $se(3)$, the adjoint operator satisfies the following similarity transformation:
\begin{equation}
    [\mathcal{V}_a] = T_{a,b} [\mathcal{V}_b] T^{-1}_{a,b}.
    \label{eq:Adjoint_end}
\end{equation}

As an example, the same twist can be expressed in a different frame. The \textit{body twist}, which is the spatial velocity expressed in the body frame, is defined as:
\begin{equation}
    [\mathcal{V}_b] = T_{s,b}^{-1} \dot{T}_{s,b} \in se(3).
\end{equation}
Once the transformation matrix $T_{s,b}$ is known, the body twist $\mathcal{V}_b$ is related to the space twist $\mathcal{V}_s$ through the adjoint transformation:
\begin{equation}
    \mathcal{V}_b = \mathrm{Ad}_{T_{b,s}}(\mathcal{V}_s),
\end{equation}
where $T_{b,s} = T_{s,b}^{-1}$. If $T_{s,b}=(R, p)$ is given as
then its inverse is computed as
\begin{equation}
    T_{b,s} = T_{s,b}^{-1} = (R^T, -R^Tp).
\end{equation}
As another example, consider Figure~\ref{fig:open_chain} (B). The screw axis of the third crease, transformed after the first and second crease rotations, becomes:
\begin{equation}
    \xi_3' = \mathrm{Ad}_{e^{[\xi_1]\theta_1}e^{[\xi_2]\theta_2}}(\xi_3).
\end{equation}

The \textit{space Jacobian} $J_s \in \mathbb{R}^{6 \times n}$ maps the joint velocity vector $\dot{\theta} \in \mathbb{R}^n$ to the space twist through the relation:
\begin{equation}
    \mathcal{V}_s = J_s \dot{\theta}.
    \label{eq:SpaceJac_start}
\end{equation}
By directly substituting the PoE formula (Equation~\ref{eq:PoE}) into the definition of the space twist (Equation~\ref{eq:space_twist}), the $i^{\text{th}}$ column of the space Jacobian is given by:
\begin{equation}
    J_{s,i} = \xi_i' = \mathrm{Ad}_{e^{[\xi_1]\theta_1} \cdots e^{[\xi_{i-1}]\theta_{i-1}}}(\xi_i).
        \label{eq:SpaceJac_end}
\end{equation}

\subsection{\label{appendix:ScrewTheory2}Rotation Formulation for Non-Perforated Loops}
\begin{figure}[!ht]
\centering
\includegraphics[width=0.45\textwidth]{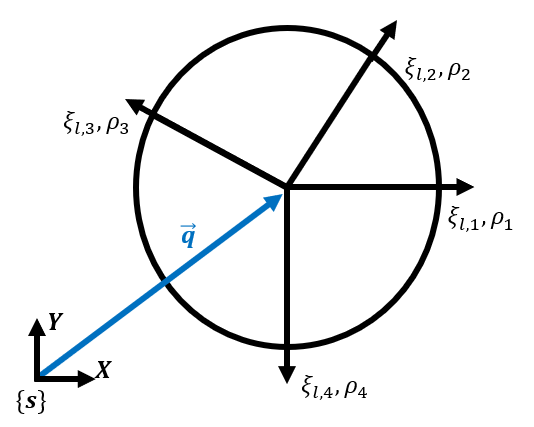}
\caption{Geometric interpretation of common $q$-point assumption in Tachi’s loop closure method}
\label{fig:tachi_method}
\end{figure}

For rigid origami without perforations, \textcite{tachi2009} proposed a loop closure constraint method using only rotational components, based on the DH convention.
In our formulation, this corresponds to a special case in which all creases in a loop intersect at a common point $q$ in the space frame, as illustrated in Figure~\ref{fig:tachi_method}.

From the $l^{\text{th}}$ loop, the $i^{\text{th}}$ column of the space Jacobian is given by:
\begin{equation}
J_{s,i}^l = \mathrm{Ad}_{T_{1:i-1}}(\xi_{l,i}),
\end{equation}
where
\begin{equation}
T_{1:i-1} = \begin{bmatrix}
R_{1:i-1} & p_{1:i-1} \\
0 & 1
\end{bmatrix},
\quad
\xi_{l,i} = \begin{bmatrix}
\omega_{l,i} \\
v_{l,i}
\end{bmatrix}.
\end{equation}
The rotation matrix $R_{1:i}$ can also be constructed recursively via the exponential map of each joint rotation:
\begin{equation}
R_{1:i} = e^{[\omega_{l,1}] \rho_1} e^{[\omega_{l,2}] \rho_2} \cdots e^{[\omega_{l,i}] \rho_i}.
\end{equation}
By definition, $v_{l,i} = -\omega_{l,i} \times q = -[\omega_{l,i}] q$, and thus the Jacobian column simplifies to:
\begin{equation}
J_{s,i}^l = \begin{bmatrix}
R_{1:i-1} \omega_{l,i} \\
[q] R_{1:i-1} \omega_{l,i}
\end{bmatrix}.
\end{equation}
This result shows that the bottom three rows (representing linear velocity) are linearly dependent on the top three rows (representing angular velocity), thereby justifying the redundancy of the translational components in the loop constraint.

Therefore, when the loop is non-perforated and all creases in the loop intersect at a common vertex $q$, the truncated loop space Jacobian $J_{s}^{l,*} \in \mathbb{R}^{3 \times n}$ can be defined as:
\begin{equation}
J_{s}^{l,*} = \begin{bmatrix}
\omega_{l,1} & R_{1:1} \omega_{l,2} & \cdots & R_{1:n_l-1} \omega_{l,n_l}
\end{bmatrix}.
\label{eq:Tachi's Jacobian}
\end{equation}
This Jacobian can be globally indexed by mapping each column to the corresponding crease variable in $\theta$, while filling the columns corresponding to non-participating creases with zero vectors.

Consequently, if the structure contains $L_o$ non-perforated (origami-type) loops and $L_k$ perforated (kirigami-type) loops, such that the total number of loops is $L = L_o + L_k$, then the full constraint matrix $A(\theta)$ has the dimension:
\begin{equation}
A(\theta) \in \mathbb{R}^{(3L_o + 6L_k) \times n}.
\end{equation}

\end{appendices}

\end{document}